\documentclass{article}

\usepackage{PRIMEarxiv}

\usepackage[utf8]{inputenc} 
\usepackage[T1]{fontenc}    
\usepackage{hyperref}       
\usepackage{url}            
\usepackage{booktabs}       
\usepackage{amsfonts}       
\usepackage{nicefrac}       
\usepackage{microtype}      
\usepackage{lipsum}
\usepackage{fancyhdr}       
\usepackage{graphicx}       
\graphicspath{{media/}}     

\usepackage{amsmath,amssymb,amsfonts}

\usepackage{graphicx}
\usepackage{textcomp}

\usepackage{ulem}
\usepackage{comment}
\usepackage{multirow}
\usepackage{multicol}
\usepackage{xcolor}

\usepackage{algorithm}
\usepackage{algpseudocode}
\usepackage{hyperref}

\title{A Survey on Reservoir Computing and its Interdisciplinary Applications Beyond Traditional Machine Learning
\thanks{\textit{\underline{Citation}}: 
\textbf{Heng Zhang and Danilo Vasconcellos Vargas. DOI:10.1109/ACCESS.2023.3299296}} 
}

\author{
  Heng Zhang, Danilo Vasconcellos Vargas \\
  Department of Information Science and Technology \\
  Kyushu University \\
  Fukuoka, Japan \\
  rogerzhangheng@gmail.com, vargas@inf.kyushu-u.ac.jp
}

\begin{document}
\maketitle

\begin{abstract}
Reservoir computing (RC), first applied to temporal signal processing, is a recurrent neural network in which neurons are randomly connected. Once initialized, the connection strengths remain unchanged. 
Such a simple structure turns RC into a non-linear dynamical system that maps low-dimensional inputs into a high-dimensional space. The model's rich dynamics, linear separability, and memory capacity then enable a simple linear readout to generate adequate responses for various applications.
RC spans areas far beyond machine learning, since it has been shown that the complex dynamics can be realized in various physical hardware implementations and biological devices. This yields greater flexibility and shorter computation time.
Moreover, the neuronal responses triggered by the model's dynamics shed light on understanding brain mechanisms that also exploit similar dynamical processes.
While the literature on RC is vast and fragmented, here we conduct a unified review of RC's recent developments from machine learning to physics, biology, and neuroscience. 
We first review the early RC models, and then survey the state-of-the-art models and their applications. 
\textcolor{black}{We further introduce studies on modeling the brain's mechanisms by RC.
Finally, we offer new perspectives on RC development, including reservoir design, coding frameworks unification, physical RC implementations, and interaction between RC, cognitive neuroscience and evolution. 
}

\end{abstract}


\section{Introduction}
\label{sec:introduction}
Artificial neural networks (ANNs) attract attention in the fields of artificial intelligence, neuroscience, computer science, and machine learning. 
These ANNs can be mainly divided into two architectures: (1) feed-forward neural networks (FFNNs) and (2) recurrent neural networks (RNNs) \cite{lecun2015deep}.
\textcolor{black}{
In the field of neuroscience, it has been realized that the convergent feed-forward circuit observed in the cerebral cortex of mammals is a method used to encode relations, allowing cognitive objects to be represented through multi-layered feed-forward architectures \cite{singer2021cerebral}.
}
In machine learning, training FFNNs is a process that usually involves the optimization of a highly non-convex problem using gradient descent based methods to find the optimum. 
One of the biggest advantages of FFNNs is their ability to deal with static (non-temporal) data processing tasks such as image recognition \cite{rawat2017deep}, object detection \cite{zhao2019object} and semantic segmentation \cite{garcia2017review}. 
However, samples are normally independently processed in FFNNs, making it hard to handle temporally correlated events without memory.

On the other hand, RNNs are models where neurons are recurrently coupled with feedback connections. 
The recurrent connections provide rich non-linear dynamics and memory, which are essential for temporal data and sequential processing.
\textcolor{black}{
However, RNNs can be challenging to train.  
This is mainly because they must deal with vanishing and exploding gradient problems, along with other problems such as longer training time and the need for careful weight initialization \cite{ribeiro2020beyond, hanin2018neural}.}
Back-propagation-through-time (BPTT) \cite{werbos1990backpropagation} and Long Short-Term Memory (LSTM) \cite{hochreiter1997long} networks are two solutions to some of the problems mentioned above. However, the learning difficulty still exists.

\begin{figure*}[ht]
    \centering
    \includegraphics[width=0.995\textwidth]{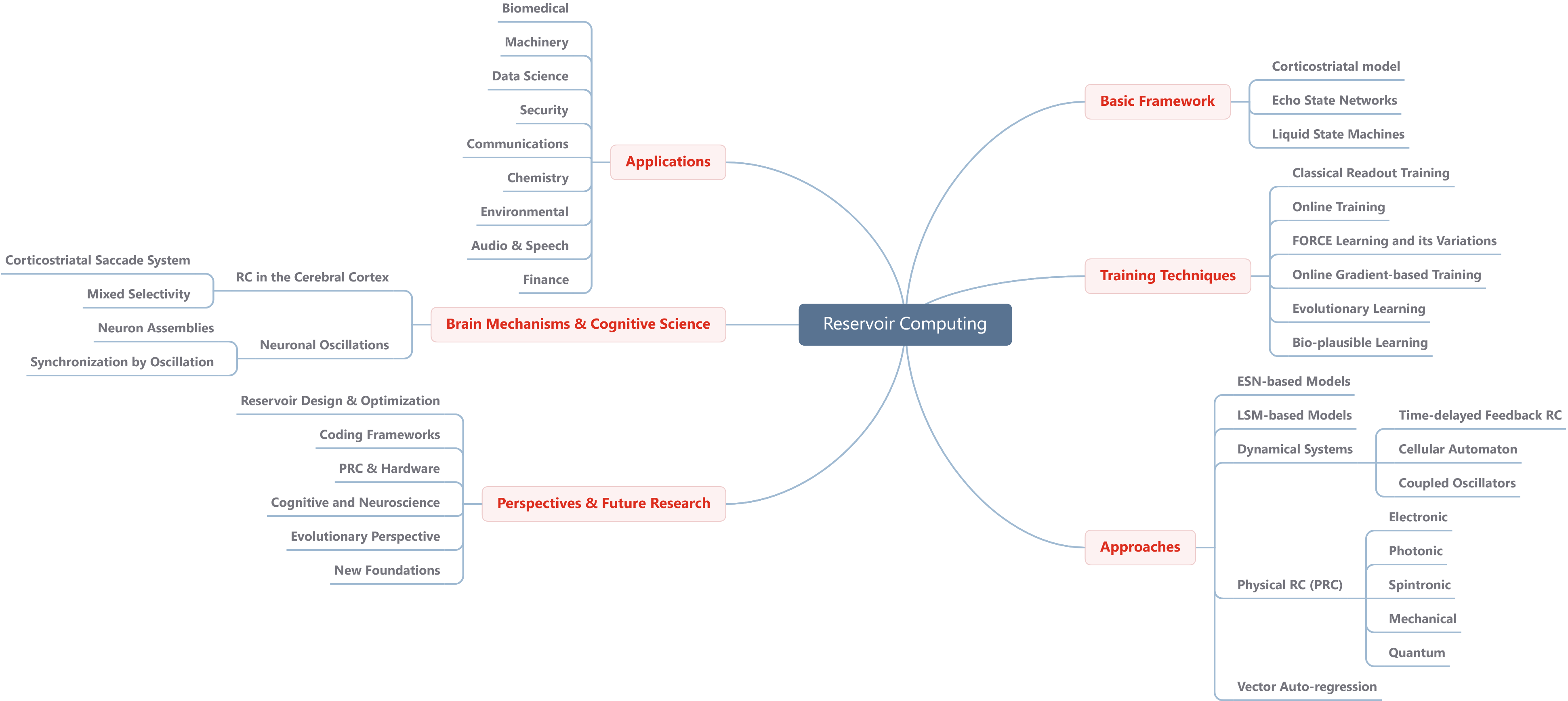}
    \caption{\textcolor{black}{Paper structure showing an overall picture and future trends of research in reservoir computing.}
    }
    \label{fig:overview}
\end{figure*} 

Reservoir computing (RC) is a bio-inspired, RNN based computational framework mainly originated from two independent models in the early 2000s: 
the Echo State Network (ESN) \cite{jaeger2001echo} and the Liquid State Machine (LSM) \cite{maass2002real}. 
Similar and related concepts include backpropagation-decorrelation (BPDC) learning algorithm \cite{steil2004backpropagation} and the cortico-striatal models realized in prefrontal cortex \cite{dominey1995complex}. 
These proposed models were then unified as the RC computational framework \textcolor{black}{by Verstraeten et, al. \cite{verstraeten2007experimental}, and were summarized in two reviews \cite{lukovsevivcius2009reservoir, schrauwen2007overview}. }

A typical RC network includes a \textit{reservoir} and a \textit{readout} layer. 
\textcolor{black}{Specifically, the reservoir is a network with randomly connected neurons. It can generate complex and high-dimensional transient responses to input. }
Such responses are considered as reservoir's states, forming unique trajectories for each input. 
These high-dimensional reservoir's states are then processed with a simple readout layer, generating the output.
During RC training, the weight connections between neurons in the reservoir are \textcolor{black}{usually} fixed to their initial random values (i.e., remain untrained), and only the readout layer is trained by relatively simple learning algorithms \textcolor{black}{(e.g., linear regression)}. 

RC has several advantages, such as a fast training process, simplicity of implementation, reduction of computational cost, and no vanishing or exploding gradient problems, among others. 
Moreover, they do have memory and, for this reason, are capable of tackling temporal and sequential problems.
These advantages have attracted increasing interest in RC research and many of its applications since its conception.
However, due to the impressive wide-range results of gradient descent based neural networks—the deep learning revolution, RC research receded into a niche for a few years after the 2010s.
\textcolor{black}{Recently, there has been a resurgence of interest in RC as a compelling biological model of neural networks, driven by its potential for scalability through physical implementations.}
Additionally, problems in deep learning such as adversarial attacks and robustness/adaptiveness issues have also contributed to an increase in interest in alternative paradigms \cite{goodfellow2014explaining, moosavi2017universal, tsipras2018robustness, su2019one, kotyan2020evolving, kotyan2022adversarial}. 

\subsection{Aims}
The complex dynamics in the reservoir indicate that RC is not only just a machine learning framework, but also a concept that highly correlates to physics, biology, and neuroscience.
In 2019, a review in physical RC (PRC) comprehensively summarizes various physical materials and hardware components that can be used for PRC implementations \cite{tanaka2019recent}. 
In biology and neuroscience, it was reported in 2013 that parts of the brain's mechanisms are similar to those of reservoirs, including mixed selectivity \cite{rigotti2013importance} and neuronal oscillations \cite{singer2021cerebral}.
\textcolor{black}{Although valuable surveys exist in specific research branches, there is still a notable gap in the literature regarding understanding the intricate interplay between RC and its connections to physics, biology, and neuroscience. The absence of such a comprehensive review hinders the advancement of our knowledge and the exploration of RC's potential across diverse domains. Addressing this need and providing a holistic overview would significantly contribute to bridging this gap, facilitating further research, and unlocking new insights into the various aspects of RC.}



\textcolor{black}{
This survey provides a clear exposition of recent developments in RC. 
While the literature on RC is vast and fragmented, we aim to provide a uniform introduction to RC.
We begin in \textbf{Section \ref{Basic framework of RC}} by introducing the fundamental concepts of RC.
In \textbf{Section \ref{Training a RC model}}, we explore different network architectures and optimization techniques that can enhance reservoir performance.
\textbf{Section \ref{Recent approaches in RC}} further highlights the recent trends of RC, facilitated by improved training schemes. Various physical hardware solutions are also reviewed, covering electronic, optical, and biophysical approaches, among others. 
Furthermore, we demonstrate the wide range of applications of RC in \textbf{Section \ref{Recent Applications of Reservoir Computing}}, including practical engineering, natural science, and social/data science, etc. These applications often involve time-dependent data, requiring memory and memory-related processing.
Next, by discussing the cortico-striatal models and coupled oscillator networks in \textbf{Section \ref{RC with brain mechanisms and cognitive science}}, we show that high-dimensional responses triggered by the reservoir's dynamics offer insights into brain mechanisms that also exploit a high-dimensional dynamical process.
Additionally, our contributions to RC reside in \textbf{Section \ref{perspectives}}, where we present fresh perspectives on the development of RC and pinpoint open problems that require further research. 
}

\section{Basic Framework of RC} \label{Basic framework of RC}
\textcolor{black}{In this section, we first review the historical developments in RC. 
By tracing its trajectory from initial theoretical foundations to the present state, our emphasis will then be on highlighting significant works that have played a crucial role in shaping the current landscape of RC.
}
\subsection{History of RC Developments}

\textbf{The first prototype of reservoir computing.} 
Perhaps the birth of the RC framework was in the 1980s-90s. 
During that period, some researchers were focusing on the characterization of the fast eye movements (i.e., the oculomotor saccade) in the corticostriatal system—the interactions between cortex and basal ganglia \cite{bruce1985primate}. 
In a pioneering work 
\textcolor{black}{(1989)}, 
Barone and Joseph \cite{barone1989prefrontal} examined the function of the corticostriatal system by carrying saccade experiments on macaque monkeys. 
\textcolor{black}{They found that some neurons have a preferred spatial saccade amplitude and direction, resulting in selective responses to a particular sequential order. 
}
This finding was further characterized as \textit{mixed selectivity} by \cite{rigotti2013importance} in 2013, which is one of the important principles in both RC and cognitive science (see Section \ref{RC with brain mechanisms and cognitive science}).

Taking inspiration from the experiments of the corticostriatal saccade system \cite{barone1989prefrontal,dominey1992cortico}, Dominey \cite{dominey1995complex,dominey1995model} developed the very first prototype of the reservoir network 
\textcolor{black}{in 1995 
(see Section \ref{RC with brain mechanisms and cognitive science} and Fig. \ref{fig:cog_PFC} for detail). }
Specifically, the model was built based on a recurrent prefrontal cortex (PFC) system (the reservoir), and a reward-related learning method in PFC-to-caudate connections (the readout). 
Since they found that the modifications of the recurrent connections are considerably computationally costing, they decided to initialize the PFC layer with a mixture of fixed inhibitory and excitatory recurrent connections (i.e., a reservoir with fixed connections between neurons). 
The reservoir was then connected to the caudate or striatum to obtain the readout. 

\textbf{Early research and problems identified in general RNNs.} 
\textcolor{black}{The main branch of RC was originated} in the fields of temporal and sequential pattern recognition using RNNs. 
In contrast to FFNNs that aim to approximate non-linear input-output \textit{functions}, RNNs are capable of representing
\textit{dynamical systems} and processing sequential inputs with recurrent connections \cite{tanaka2019recent,lukovsevivcius2009reservoir}. 

Early studies of RNNs include a well-known model called Hopfield network \cite{hopfield1982neural} 
\textcolor{black}{in the 1980s}. 
The network topologies were specifically formulated with symmetrical weights connections and were trained in unsupervised ways. 
This special type of network normally experiences chaotic or stochastic dynamics with the mathematical background of statistical physics \cite{lukovsevivcius2009reservoir}.  
Another type of RNN features a deterministic update dynamics and directed weighted connections. Systems from this type of RNN are usually made of high dimensional hidden states with non-linear dynamics, resulting in a transformation from an input sequence into an output sequence.
Two standard examples are (1) back-propagation-through-time (BPTT) by \cite{werbos1990backpropagation,rumelhart1985learning}; and (2) real-time recurrent learning (RTRL) by \cite{williams1989learning}. 
Even though these learning methods showed great potential in complex sequential processing, they struggled to tackle real-world problems due to the high computational costs and difficulties of training, especially the vanishing and exploding gradient problems that make them hard to capture long-term dependencies \cite{ribeiro2020beyond, hanin2018neural}.
\textcolor{black}{In 1997}, 
a well-known architecture, Long Short Term Memory (LSTM) \cite{hochreiter1997long}, was then proposed to address these problems. For more details on the gradient-based RNNs, please refer to an early review \cite{atiya2000new}. 

\textbf{Unification of reservoir computing.} 
\textcolor{black}{In 2000}, 
\cite{atiya2000new} proposed a new algorithm based on error gradient approximation, which efficiently reduces the computational complexity and shows faster convergence in recurrent network training. 
This work, referred to Atiya and Parlos Recurrent Learning (APRL) in later literature, identified that an RNN can be divided into two parts: 
the quickly changing output weights, and the slowly adapting hidden weights.
Therefore, APRL is considered the algorithm to bridge between general RNNs and reservoir computing \cite{lukovsevivcius2009reservoir}. Besides, another predecessor of RC, the backpropagation-decorrelation (BPDC) learning algorithm, further simplified APRL and made it an online learning algorithm \cite{steil2004backpropagation}.

\textcolor{black}{Later in the early 2000s}, two types of fundamental reservoir computing algorithms were independently invented by Maass et al. \cite{maass2002real} as Liquid State Machine (LSM), and by Jaeger \cite{jaeger2001echo} as Echo State Network (ESN). The two algorithms, as well as other related works such as BPDC and works in neuroscience fields such as Dominey's research \cite{dominey1995complex}, were unified as a computational framework called \textit{reservoir computing} (RC) \cite{verstraeten2007experimental}. 
In this unified framework, the low-dimensional input data is transformed into spatio-temporal patterns in a high-dimensional space by the \textit{reservoir}—an RNN with fixed topologies and unchanged weights. 
The high-dimensional responses generated by the reservoir are then processed by the \textit{readout}—an output layer which can be trained with simple learning algorithms such as linear regression. 
In other words, during training, the values of the weight connections within the reservoir remain unchanged while only the readout weights are trained based on specific tasks. 
Before going deep into recent advances in RC, we will first introduce the basic concepts of ESN and LSM in the following subsections. 

\subsection{Echo State Networks}
\begin{figure}[ht]
    \centering
    \includegraphics[width=0.6\linewidth]{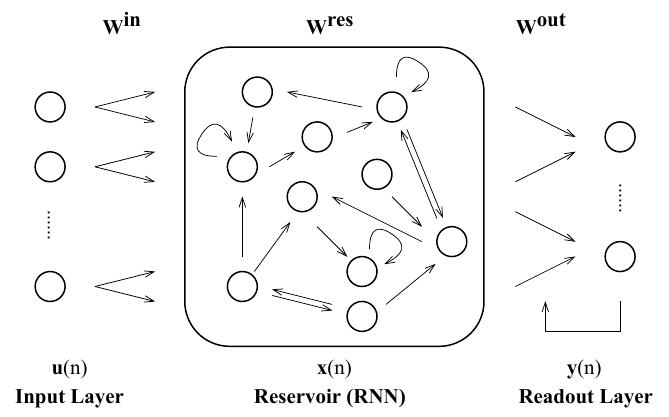}
    \caption{Simplified structure of Echo State Network (ESN).
    }
    \label{fig:ESN archi}
\end{figure} 

Echo State Network (ESN) was first proposed by \cite{jaeger2001echo}. This pioneering work is based on the fact that training only the readout layer of an RNN can achieve acceptable performance, if the network has sufficiently rich dynamics. 
\textcolor{black}{
ESN is normally implemented with leaky-integrated, non-spiking, discrete-time and continuous-value artificial neurons (see Fig. \ref{fig:ESN archi} the network structure).
To illustrate the technical details, here we use the notations by \cite{lukovsevivcius2012practical}. 
Consider a temporal processing task, where the input signal is $\mathbf{u}(n) \in \mathbb{R}^{N_{u}}$ and the desired target signal is $\mathbf{y}^{target}(n) \in \mathbb{R}^{N_{y}}$, given $n=1,...,T$ with $T$ being the total number of discrete data points. 
The goal is to generate an output signal $\mathbf{y}(n) \in \mathbb{R}^{N_{y}}$ that matches $\mathbf{y}^{target}(n)$ as optimally as possible by minimizing the error between the two signals (e.g., Mean-Square Error, MSE).
The simplified update equations of the reservoir part in ESNs are given by:
}

\textcolor{black}{
\begin{equation}
    \label{eq:esn1}
    \mathbf{\Tilde{x}}(n)=tanh(\mathbf{W}^{in}\mathbf{u}(n)+
    \mathbf{W}\mathbf{x}(n-1)),
\end{equation}
\begin{equation}
    \label{eq:esn2}
    \mathbf{x}(n)=(1-\alpha)\mathbf{x}(n-1) + \alpha\mathbf{\Tilde{x}}(n),
\end{equation}
}

\noindent where 
\textcolor{black}{
$\mathbf{\Tilde{x}}(n) \in \mathbb{R}^{N_{x}}$ is the update at time step $n$,
$\mathbf{x}(n) \in \mathbb{R}^{N_{x}}$ is the state vector of the reservoir neurons (also known as the resulting states or the \textit{echo} of its input history \cite{lukovsevivcius2009reservoir}), 
$\mathbf{W}^{in} \in \mathbb{R}^{N_{x} \times N_{u}}$ and $\mathbf{W} \in \mathbb{R}^{N_{x} \times N_{x}}$ are the weight matrices of the input-reservoir connections and the recurrent connections inside the reservoir, respectively. $tanh()$ is the non-linear activation function applied element-wise. $\alpha$ is the leaking rate that mainly controls the speed of the dynamics. 
}

The readout layer is normally linearly defined as:
\begin{equation}
    \label{eq:esn3}
    \mathbf{y}(n)=\mathbf{W}^{out}\mathbf{x}(n),
\end{equation}

\noindent where
\textcolor{black}{$\mathbf{y}(n) \in \mathbb{R}^{N_{y}}$ is the output vector and $\mathbf{W}^{out} \in \mathbb{R}^{N_{y} \times N_{x}}$ is the weight matrix of the reservoir-readout connections.
Alternatively, one can also introduce a bias value in both reservoir and readout, as well as integrate the input signal directly to the readout layer. In this case, $\mathbf{u}(n)$ in Eq. \ref{eq:esn1} becomes $[1;\mathbf{u}(n)]$ and $\mathbf{x}(n)$ in Eq. \ref{eq:esn3} becomes $[1;\mathbf{u}(n);\mathbf{x}(n)]$, where $[\cdot;\cdot]$ represents concatenation. A brief training procedure is shown in Alg. \ref{alg1}.
}

\textcolor{black}{
\begin{center}
\scalebox{0.95}{
\begin{minipage}{1\linewidth}
\begin{algorithm}[H]
\caption{Simplified procedure of ESN training.} \label{alg1}
\begin{algorithmic}[1]
\State \textcolor{black}{Initialize the network by generating random $\mathbf{W}^{in}$, $\mathbf{W}$. It is common to use uniform distributed randomization $U(-1,1)$.}
\State \textcolor{black}{Run the model with input signal $\mathbf{u}(n)$, $n=1,...T$, it will generate the same length of the reservoir states $\mathbf{x}(n)$ by Eq. \ref{eq:esn1}-\ref{eq:esn2}.}
\State \textcolor{black}{Collect all $\mathbf{x}(n)$, then calculate and get the output signal $\mathbf{y}(n)$ by Eq. \ref{eq:esn3}.}
\State \textcolor{black}{Minimize the MSE between $\mathbf{y}(n)$ 
 and $\mathbf{y}^{target}(n)$ using techniques such as linear regression. This should obtain a well-trained $\mathbf{W}^{out}$.}
\State \textcolor{black}{Take unseen data $\mathbf{u}_{test}(n)$ and obtain the predicted and/or generated output $\mathbf{y}_{test}(n)$.}
\end{algorithmic}
\end{algorithm}
\end{minipage}
}
\end{center}
}

During conventional ESNs training, $\mathbf{W}^{in}$ and $\mathbf{W}$ remain unchanged, and only $\mathbf{W}^{out}$ is trained in order to minimize the error between the network output and the target output (teacher signal), usually by using linear regression such as ridge regression \cite{hoerl1970ridge}.
Alternatively, many new proposed learning rules for ESN training exist,
including but not limited to online FORCE learning \cite{sussillo2009generating}, weights pre-training \cite{zhong2017genetic}, gradient-based training \cite{thiede2019gradient}, and evolutionary learning \cite{wang2015optimizing}. 
For details on training a RC model, please refer to section \ref{Training a RC model}.

\textbf{Echo state property.} 
An essential condition (requisite) that a standard ESN must meet is the echo state property (ESP), which ensures a condition of asymptotic state convergence of the reservoir.
This property is under the influence of both the reservoir and the given input. 
On one hand, ESP is an algebraic property that is controlled by the reservoir's weight matrix $\mathbf{W}$.
It has been mathematically analyzed in \cite{lukovsevivcius2009reservoir, yildiz2012re, lukovsevivcius2012practical} that the spectral radius ($SR$, i.e., the maximum eigenvalue of $\mathbf{W}$) smaller than unity ensures ESP in most situations. As a result, many RNN-based RCs in literature consider $SR<1$ as a necessary condition to make the models work (see the introduction section in \cite{schrauwen2007overview}). However, it had been proved in \cite{yildiz2012re} that $SR<1$ is neither sufficient nor necessary for the ESP. The author claim that 
\textcolor{black}{
``it is not required to scale the spectral radius below 1, and there is no general benefit in scaling the spectral radius toward the Edge of Chaos''.} The same paper also proposed new sufficient conditions for the ESP. Please refer to \cite{yildiz2012re} for mathematical details.
On the other hand, ESP has been empirically studied in the presence of driving inputs of varied strength 
\cite{gallicchio2018chasing}, without looking at the mathematics. \cite{gallicchio2018chasing} shows that for an input-driven reservoir and a proper input scaling, the actual range of ESP validity (i.e., $SR$), is much wider than what is covered by the above literature conditions.

\textbf{Memory capacity and edge of chaos.}
As a special type of RNNs, ESN also has the characteristic of short-term memory.
Analytical results that characterize the dynamical short-term memory capacity of reservoirs were discussed in \cite{lukovsevicius2012reservoir, jaeger2002short}. 
Meanwhile, it can be found in a good deal of literature that reservoirs are claimed to work best when they are tuned to operate at the so-called ``edge of chaos'' \cite{bertschinger2004real, legenstein2007edge}. Here, the edge of chaos refers to a region of parameter settings which makes the dynamical system operates at the boundary between the chaotic and non-chaotic behavior. However, this is also a misnomer, as claimed by Jaeger in \cite{nakajima2021reservoir, yildiz2012re} that the ``edge'' in question here is the edge of the ESP, not the edge of chaos. For a detailed discussion, please refer to \cite{cramer2020control}.

\subsection{Liquid State Machines}
\begin{figure}[ht]
    \centering
    \includegraphics[width=0.6\linewidth]{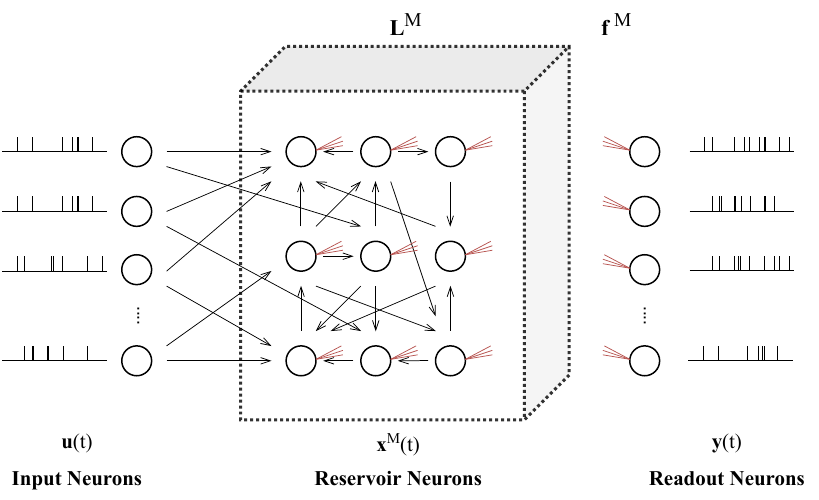}
    \caption{Simplified structure of Liquid State Machine (LSM). Note that LSM is implemented by spiking integrate-and-fire neurons, rather than non-spiking artificial neurons used in ESN.
    }
    \label{fig:LSM archi}
\end{figure} 

Liquid State Machine (LSM) was proposed by \cite{maass2002real} from the perspective of computational neuroscience in order to study the brain mechanisms and model the neural microcircuits. 
In contrast to ESN that uses non-spiking artificial neurons, LSM is more biologically plausible as it is based on the Spiking Neural Networks (SNNs) with recurrent reservoir structures.
Inside the reservoir, usually a 3D structured and locally connected network of spiking integrate-and-fire (IF) neurons is randomly created and stimulated by external input spike train signals (see Fig. \ref{fig:LSM archi}). Intuitively, the reservoir in LSM is often called liquid,
since they follow a metaphor of excited states as ripples on the surface of a pool of water \cite{maass2002real}.
\textcolor{black}{
Similar to ESNs in its form, here we use $\mathbf{u}(t)$, $\mathbf{x}(t)$ and $\mathbf{y}(t)$ to represent the input, reservoir state and output, respectively.
The reservoir dynamic of LSM is given by}:

\begin{equation}
    \label{eq3}
    \mathbf{x^M}(t)=\mathbf{L^M u}(t),
\end{equation}

\noindent where $t$ represents continuous time, $\mathbf{x^M}$ is the reservoir state, $\mathbf{u}$ represents the input spikes, and $\mathbf{L^M}$ is the \textit{liquid filter} for input-reservoir state transformation. The readout is given by:

\begin{equation}
    \label{eq4}
    \mathbf{y}(t)=\mathbf{f^M}(\mathbf{x^M}(t)),
\end{equation}

\noindent where $\mathbf{y}(t)$ is the output vector and $\mathbf{f^M}$ is a ``memory-less'' \textit{readout map}. The readout here can also be trained using simple algorithms.

\textbf{Separation and approximation properties.} LSMs have two mathematical preconditions, namely separation property (SP) and approximation property (AP). 
These two properties ensure that the network has fading memory (i.e., echo state property in ESN). 
Specifically, SP addresses the degree of separation between different internal states $\mathbf{x}$ caused by different input $\mathbf{u}$ (condition is met if the \textit{liquid filter} $\mathbf{L^M}$ satisfies the point-wise separation property), 
whereas AP addresses the capability of the readout layer to produce the target outputs given different liquid states $\mathbf{x}$ (condition is met if the \textit{readout map} $\mathbf{f^M}$ satisfies the approximation property). 
For the mathematical basis of SP and AP, please refer to \cite{maass2002real}. Overall, these two properties in LSMs, together with the ESP and memory capacity in ESNs literature, ensure the RNN-based reservoirs function properly.

\subsection{Comparison of ESNs \& LSMs}
ESNs and LSMs are two similar RNNs with reservoir structures. The main difference between them is that LSM used spiking IF neurons, while ESN is based on non-spiking neurons. 
This makes LSMs more biologically plausible to be used in investigating biological mechanisms of information processing in the brain.
In terms of model implementation, although both models show noticeable advantages in reducing computation cost and training time, LSM, with its biologically inspired characteristics and spiking implementation, becomes more suitable for new types of hardware
such as neuromorphic chips \cite{zhang2015digital}. Therefore, LSM is reported with several hardware designs and applications.

ESNs show better flexibility in model modifications, as many variants of ESNs were proposed to enhance the network performances (see Section \ref{Training a RC model}). 
These modifications are mainly to overcome the disadvantages of conventional ESNs. The first drawback is that the fixed connection weights could limit the performance of ESNs, since they are randomly initialized without the process of tuning or optimizing. 
Moreover, it is shown that the improvement of ESNs performance will reach a saturation as the size of the reservoir increases to a certain amount. This means that only increasing reservoir size may not result in a better performance. 
As a result, researchers have been focusing on constructing multi-reservoir ESNs, such as multi-layered ESNs \cite{zhang2019deep, gallicchio2017deep, long2019evolving} and parallel reservoir computing \cite{alomar2020efficient}. 
Meanwhile, several optimization approaches have been proposed to fine-tune the networks weights and hyperparameters by using either evolutionary algorithms \cite{long2019evolving, chouikhi2017pso} or gradient based optimization techniques \cite{thiede2019gradient}.
Detailed discussions of recent ESN and LSM models are covered in section \ref{Recent approaches in RC}.


\section{Training a RC Model} \label{Training a RC model}

\begin{table*}[]
\centering
\caption{Various training techniques of RC.}

\renewcommand{\arraystretch}{1.0}
\resizebox{0.9\textwidth}{!}{
\centering
\begin{tabular}{lllll}
\hline \hline
 & \textbf{Category of techniques}                &  & \textbf{Training approaches}                                                          &  \\ \cline{2-2} \cline{4-4}
 & Classic readout training &  & \begin{tabular}[c]{@{}l@{}}Linear regression\\ Ridge regression \cite{hoerl1970ridge} \end{tabular} &  \\ \cline{2-2} \cline{4-4}
 &
  Online learning techniques &
   &
  \begin{tabular}[c]{@{}l@{}}Least mean squares (LMS) \cite{jaeger2002adaptive, jaeger2004harnessing} \\ Recursive least squares (RLS) \cite{tamura2021transfer} \\ FORCE learning \cite{sussillo2009generating} \\ FORCE variations \cite{triefenbach2010phoneme, sussillo2012transferring, laje2013robust, depasquale2018full, nicola2017supervised, tamura2020two, tamura2021transfer, zheng2020r}\end{tabular} &
   \\ \cline{2-2} \cline{4-4}
 &
  Online gradient based training &
   &
  \begin{tabular}[c]{@{}l@{}}LSM method \cite{jaeger2002adaptive, jaeger2004harnessing} \\ GD-based optimizer \cite{thiede2019gradient} \\ BackPropagation-DeCorrelation (BPDC) \cite{lukovsevivcius2009reservoir} \\ Back-propagation Through Time (BPTT)\\ ACTRNN  \cite{heinrich2020learning}\end{tabular} &
   \\ \cline{2-2} \cline{4-4}
 &
  Evolutionary learning techniques &
   &
  \begin{tabular}[c]{@{}l@{}}Genetic algorithm (GA) \cite{sharma2017optimized} \\ Particle swarm optimization (PSO) \cite{basterrech2014experimental, chouikhi2017pso} \\ Competitive swarm optimizer (CSO) \cite{long2019evolving} \\ Evolino \cite{schmidhuber2007training} \end{tabular} &
   \\ \cline{2-2} \cline{4-4}
 &
  Biologically plausible learning techniques &
   &
  \begin{tabular}[c]{@{}l@{}}Hebbian learning \cite{jaeger2005reservoir} \\ Spike Timing Dependent Plasticity (STDP) \cite{norton2006preparing, jin2016sso, luo2018improving} \\ Intrinsic Plasticity (IP) \cite{steil2007online, schrauwen2008improving, xue2017reservoir, li2011model} \end{tabular} &
   \\ \hline \hline
\end{tabular}
}
\label{table: RC training}
\end{table*}

\textcolor{black}{
In this section, we will delve into an array of methodologies for training and optimizing RC models (see Table \ref{table: RC training}). Beyond simply training the readout, we will explore diverse techniques that aspire to enhance the construction of reservoirs from various aspects, thereby enabling more effective and efficient models.
}
These include but does not limit to (1) classical readout training such as ridge regression \cite{hoerl1970ridge}; 
(2) online learning such as least mean square method and FORCE learning \cite{sussillo2009generating};
(3) pre-training such as particle swarm optimization \cite{chouikhi2017pso};
(4) online gradient based learning with back-propagation \cite{bellec2019biologically, bellec2020solution};
(5) evolutionary learning such as Evolino \cite{schmidhuber2007training};
and (6) biologically plausible learning techniques such as Hebbian learning \cite{hebb2005organization}.

\subsection{Classical Readout Training}
The original works of ESN and LSM state that the readout of a reservoir with rich dynamics can be trained by using any statistical classification or regression methods \cite{schrauwen2007overview}. 
\textcolor{black}{Using ESNs for example, it is recommended to apply simple linear regression technique to single-layer readout. Again, we use the notations by \cite{lukovsevivcius2012practical} for illustration. First, notice that Eq. \ref{eq:esn3} can be rewritten and extended in a matrix form as:
}

\textcolor{black}{
\begin{equation}
    \label{eq:esn4}
    \mathbf{Y} = \mathbf{W}^{out}\mathbf{X} \approx \mathbf{Y}^{target},
\end{equation}
}

\noindent where 
\textcolor{black}{$\mathbf{Y} \in \mathbb{R}^{N_y \times T}$ stands for the collection of all $\mathbf{y}(n)$. Similar notation goes to $\mathbf{X}$ and $\mathbf{Y}^{target}$. It is clear that $\mathbf{W}^{out}$ needs to be optimized to minimize the difference between $\mathbf{Y}$ and $\mathbf{Y}^{target}$. The most common technique is the ridge regression \cite{hoerl1970ridge}, where $\mathbf{W}^{out}$ is obtained by: 
}

\textcolor{black}{
\begin{equation}
    \label{eq:esn5}
    \mathbf{W}^{out} = \mathbf{Y}^{target}\mathbf{X}^{T}{(\mathbf{X}\mathbf{X}^{T} + \beta\mathbf{I})}^{-1},
\end{equation}
}

\noindent where $\mathbf{I}$ is the identity matrix with $\beta$ being the regularization factor. Detailed implementation can be found in a practical guide of ESN training in \cite{lukovsevivcius2012practical}. Once $\mathbf{X}$ is obtained in an off-line way, one can tune $\beta$ to reach the best performance without any model retraining.
Ridge regression often shows sufficiency in many concrete tasks, when the reservoir provides rich non-linear dynamics. 
Moreover, it is easy and fast to train, which attracts many researchers coming from non-machine learning backgrounds. 

\subsection{Online Learning Techniques}
\cite{lukovsevivcius2012practical} provides many empirical solutions on how to produce a reservoir with good initialization. However, problems still exist, as one cannot guarantee that the reservoir is always well-initialized. Another effective way to solve this problem is online adaptation. In this manner, a feedback loop between the reservoir and readout is usually introduced. 

\subsubsection{Least Mean Squares (LMS) and Recursive Least Squares (RLS) Methods} 
Originated in the area of adaptive signal processing, LMS and RLS are the two standard online learning methods for reservoir computing models \cite{farhang2013adaptive}. 
The mathematical description of LMS and RLS are presented in \cite{jaeger2002adaptive, tamura2021transfer}. 
To illustrate, LMS is a gradient-based error minimization method in which an error is exponentially discounted propagating back through time, yet this method might be unstable because of the large eigenvalue spreads of the cross-correlation matrix (i.e., $\mathbf{X}\mathbf{X}^{T}$). 
Moreover, it is reported that LMS struggles to capture the history-dependent temporal data \cite{beer2019one}. Compared to LMS, RLS is more popular due to its robustness/insensitivity to the effect of eigenvalue spreads mentioned above, as well as its faster second-order convergence speed. 
Therefore, RLS method for RC has been widely studied in \cite{nicola2017supervised, depasquale2018full, tamura2021transfer}. 
Although RLS has advantages over LMS, it is more computationally costly with $O(N^2)$ time complexity, while LMS only requires $O(N)$ in most situations \cite{lukovsevicius2012reservoir}, where $N$ is the number of variables. 
Having said that, RLS is employed by FORCE learning, which creates a new branch of reservoir computing with regard to cognitive science and brain mechanisms. 

\begin{figure}[ht]
    \centering
    \includegraphics[width=0.6\linewidth]{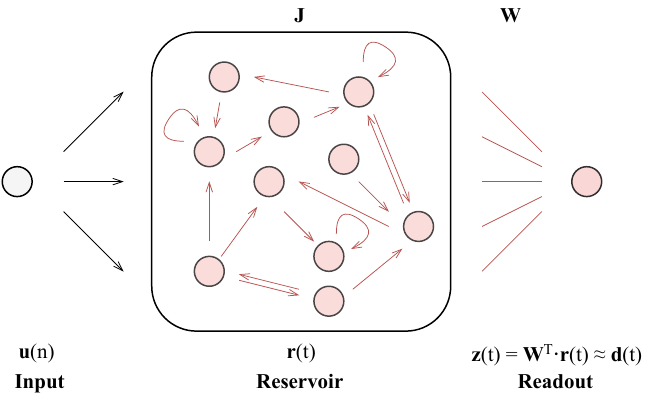}
    \caption{FORCE learning overview. The red color represents that the weight connections can be modified during training.
    }
    \label{fig:FORCE}
\end{figure} 

\subsubsection{FORCE Learning} 
It has been shown that RNNs often experience spontaneous chaotic activity, and algorithms such as BPTT \cite{rumelhart1985learning} are usually not able to converge if the network exhibits chaotic activity. ESN models address the chaotic activity by ensuring echo state property (ESP), so that the models do not operate in a chaotic manner. 
Instead of avoiding spontaneous activity like ESNs, \textit{Fisrt-Order Reduced and Controlled Error} (FORCE) learning, which is perhaps the most popular online learning method of RC, reveals that the results are better when the reservoirs exhibit chaotic behavior before training \cite{sussillo2009generating}.  
By modifying the synaptic strengths of the reservoir (either internal or external), models trained with FORCE learning show the effectiveness of suppressing autonomous chaotic activity while turning it into a wide variety of desired output patterns.
\textcolor{black}{
Since the original mathematics are quite complex, here we aim to provide a simplified description of the online readout training for completeness \cite{duane2017force}. Please refer to \cite{sussillo2009generating} for details. 
}

\textcolor{black}{
Consider the reservoir state as $\mathbf{r}(t)$ (i.e., $\mathbf{x}(n)$ in ESNs), the network output $\mathbf{z}(t)$ (i.e., $\mathbf{y}(n)$ in ESNs) is defined as:
}

\textcolor{black}{
\begin{equation}
    \label{eq:force1}
    \mathbf{z}(t) = \mathbf{w}^{T}\mathbf{r}(t),
\end{equation}
}

\noindent where 
\textcolor{black}{$\mathbf{w}$ is the weights connecting reservoir and readout. Note that here the output dimension is restricted to one, while it can be easily generalized to multidimensional. Training of $\mathbf{w}$ happens at every time interval $\Delta t$. Before updating at time $t$, the error is denoted by:
}

\textcolor{black}{
\begin{equation}
    \label{eq:force2}
    {e}_{-}(t) = \mathbf{w}^{T}(t-\Delta t)\mathbf{r}(t) - f(t),
\end{equation}
}

\noindent where 
\textcolor{black}{$f(t)$ is the predefined target function (i.e., $\mathbf{y}^{target}(n)$ in ESNs). The FORCE algorithm uses a modified RLS method to update the weights by:
}

\textcolor{black}{
\begin{equation}
    \label{eq:force3}
    \mathbf{w}(t) = \mathbf{w}(t-\Delta t) - {e}_{-}(t)\mathbf{P}(t)\mathbf{r}(t),
\end{equation}
}

\noindent where 
\textcolor{black}{$\mathbf{P}(t)$ is a square matrix that is updated at the same time as the weights according to
}

\textcolor{black}{
\begin{equation}
    \label{eq:force4}
    \mathbf{P}(t) = \mathbf{P}(t-\Delta t) - \frac{\mathbf{P}(t-\Delta t)\mathbf{r}(t)\mathbf{r}^{T}(t)\mathbf{P}(t-\Delta t)}{1+\mathbf{r}^{T}(t)\mathbf{P}(t-\Delta t)\mathbf{r}(t)},
\end{equation}
}

\noindent and
\textcolor{black}{is initialized as:
}

\textcolor{black}{
\begin{equation}
    \label{eq:force5}
    \mathbf{P}(0) = \frac{\mathbf{I}}{\alpha},
\end{equation}
}

\noindent where 
\textcolor{black}{$\mathbf{I}$ is the identity matrix with $\alpha$ as constant. After training, the error becomes
}

\textcolor{black}{
\begin{equation}
    \label{eq:force6}
    {e}_{+}(t) = {e}_{-}(t)(1-\mathbf{r}^{T}(t)\mathbf{P}(t)\mathbf{r}(t)).
\end{equation}
}

\noindent Finally, 
\textcolor{black}{the training will end when it reaches
}

\textcolor{black}{
\begin{equation}
    \label{eq:force7}
     \frac{{e}_{+}(t)}{{e}_{-}(t)} \approx 1.
\end{equation}
}

\textcolor{black}{The above modified RLS method is} applied to suppress the output errors and frequently adapt the weight matrices in the reservoir or readout (see Fig. \ref{fig:FORCE}). 
This makes FORCE learning disparate from other traditional iterative training schemes—the errors
\textcolor{black}{in FORCE learning}
are always small during training, even at the beginning, suggesting that the aim is not to reduce errors but rather to keep the errors small. When training is done, the network will autonomously generate the desired output. 
As the author claimed, FORCE learning helps to construct machine learning based RNNs that ``generate complex and controllable patterns of activity either in the absence of or in response to input''. 
\textcolor{black}{It provides an interesting link between computational and biological neuroscience. In short, FORCE learning can be seen as a useful tool for optimizing RC, and simultaneously, it presents a potential model that could help understand biological neural circuits \cite{sussillo2009generating}.
}

\begin{figure}[ht]
    \centering
    \includegraphics[width=0.7\linewidth]{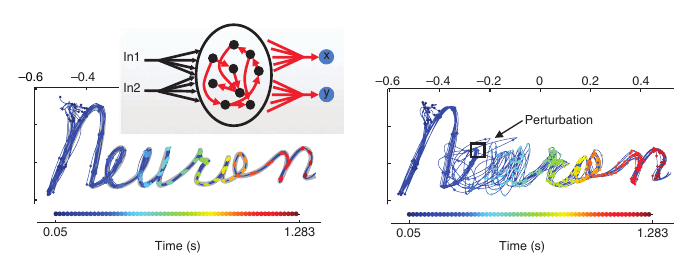}
    \caption{Generation and stability of complex spatio-temporal motor patterns by FORCE learning (modified from \cite{laje2013robust}). Left panel shows the ten trials to re-generate the target signal (in gray line) after training. Right panel shows the same trials under the addition of a perturbation. 
    }
    \label{fig:laje2013robust}
\end{figure} 

\subsubsection{FORCE Learning Variations and Implementations} 
One of the disadvantages of FORCE learning is that the trained network is too complex to analyze the neuron activities. Also, for complex real-world problems such as speech recognition, networks trained by FORCE require many more units to match the performance of gradient based networks, as reported in  \cite{triefenbach2010phoneme}.

Therefore, some studies have been published aiming at improving FORCE learning,
ranging from neuroscience (e.g., spiking networks) to physical and hardware implementations.  
\textcolor{black}{An ``extended'' FORCE \cite{sussillo2012transferring}, for example, was proposed for more general internal learning by using the desired output to generate targets for every internal neuron in the network. 
}
This so-called ``target-generating'' network is then improved by \cite{laje2013robust}, in which the reservoir network can preserve time information and generate complex and high-dimensional trajectories even under high levels of noise (see Fig. \ref{fig:laje2013robust}).  
\textcolor{black}{To this end, the FORCE variations are still infeasible to be implemented as a spiking network. 
Later, \cite{depasquale2018full} proposed a full-FORCE algorithm. 
Compared with FORCE learning, it requires fewer neurons, achieves significantly better performance in noisy environments, and can also be applied to SNNs (see implementation in \cite{nicola2017supervised}). 
For hands-on implementation, the full-FORCE has been realized using a Python spiking network framework called Nengo \cite{bekolay2014nengo}.}
Other recent improvements of FORCE learning include (1) Two-Step FORCE that converges faster than the original work \cite{tamura2020two}; 
(2) Transfer-FORCE learning which takes the advantages of both LMS and RLS methods for better learning performance \cite{tamura2021transfer}; (3) R-FORCE that aims to model multidimensional sequences \cite{zheng2020r}. 
\textcolor{black}{
Very recently, \cite{liu2022tension} built an object-oriented, open-source Python package that implements a TensorFlow / Keras API for FORCE.
}

\subsection{Online Gradient Based Training}
Reservoir based on non-spiking artificial neurons (e.g., ESNs) can be trained by using gradient descent (GD) approaches. The LMS method mentioned earlier is one of the candidates which is first proposed in \cite{jaeger2004harnessing}, yet using such approach shows poor stability even when trying to stabilize the network by adding noise. \cite{thiede2019gradient} proposed a more stable version of GD-based reservoir to optimize four hyperparameters: the input scaling, spectral radius, leaking rate, and regularization parameter.
Besides, BackPropagation-DeCorrelation (BPDC) algorithm is another powerful method for online training of single-layer readouts with feedback connected back to reservoirs. This algorithm is robust to the random initialization of the reservoir weights, and it is also capable of tracking quickly changing signals.
Detailed discussions of BPDC are presented in a survey \cite{lukovsevivcius2009reservoir}. 
Meanwhile, the classical Back-propagation Through Time (BPTT) approach for RNN training can also be applied to RC model. It is worth noting that a network architecture called Adaptive Continuous Time Recurrent Neural Network (ACTRNN) \cite{heinrich2020learning} shows some similarity to GD-based RC. 
\textcolor{black}{Please refer to Section \ref{RC with brain mechanisms and cognitive science} for details.}

\subsection{Evolutionary Learning Techniques}
\textbf{Evolutionary algorithm.} 
One of the disadvantages of the traditional RC models is that the performance is highly reliant on the random initialization of the weights and hyperparameters. 
While the optimal hyperparameters can be found by grid-search techniques, using such techniques to find the optimal weights' initialization is nearly infeasible \cite{lukovsevivcius2012practical}, given a concrete task. 
Therefore, instead of applying online learning rules, another possible direction, taking inspiration from above, is to train (or pre-train) the reservoirs using evolutionary algorithms (EA). When EAs are applied to evolve any type of neural networks (including reservoirs) they usually receive the name of neuroevolution.

Various types of EA can be used to evolve a reservoir, including (1) genetic algorithms (GA) \cite{sharma2017optimized}; (2) particle swarm optimization (PSO) \cite{basterrech2014experimental, chouikhi2017pso} and its variants \cite{long2019evolving}; and (3) artificial bee colony \cite{badem2017new}. 
For example, a GA was applied to a double-reservoir ESN for parameter optimization, yet without optimizing weights of input and the reservoirs \cite{zhong2017genetic}. 
Inspired by LSTM \cite{hochreiter1997long}, another EA for RC was proposed called
Evolino \cite{schmidhuber2007training}.
Evolino constructs units that are capable of preserving memory for long periods of time, in which the weights of the reservoir are trained using evolutionary methods. 
A performance comparison of several EAs for RC are presented in \cite{ferreira2011comparing}.

Particle swarm optimization (PSO), which is an efficient and widely used technique for finding optimal regions on complex spaces, has also applied to reservoir weight optimization. 
The first two attempts of using PSO technique include using a binary PSO to find the optimal reservoir-readout connections \cite{wang2015optimizing}, as well as a supervised PSO algorithm by \cite{basterrech2014experimental} for better initializing the input weights of RC. 
However, only a subset of the weights was tuned in the latter model, due to the high computational cost.
\cite{chouikhi2017pso} further developed the PSO algorithm for RC, where a portion of fixed weights in an ESN is pre-trained via PSO. 
The results show improvements on model generation as well as a faster convergence time, yet the network architecture is rather simple, and some hyperparameters should be selected empirically and carefully. 
The latest version of PSO based RC is the competitive swarm optimizer (CSO) for fault diagnosis problems \cite{long2019evolving}, which is a hybrid evolutionary algorithm combining both a variant of PSO and local search (LS).

\subsection{Biologically Plausible Learning Techniques}
\textbf{Hebbian learning and Spike-timing-dependent plasticity.}
Reservoir computing takes inspiration partially from biological systems. LSM-based reservoirs, for example, are implemented using spiking neurons. This indicates that some biologically plausible adaptation methods can be applied for reservoir training. 
Inspired by synaptic plasticity in human brains, the first attempts would be to use Hebbian and anti-Hebbian learning to try to decrease the eigenvalue spread in ESNs but failed \cite{jaeger2005reservoir}.
Later, it is reported in \cite{lukovsevicius2012reservoir} that the reservoir trained by Hebbian learning ``makes neurons prefer inputs that are easy to predict and weaken connections from those that carry more information''.
In terms of LSM, the spike-time-dependent plasticity (STDP), which is based on Hebbian learning and is often integrated with SNNs, 
is reported to improve the separation property (SP) in some real-world speech data \cite{norton2006preparing}. STDP was further developed to reduce memory storage load to make RC more hardware-friendly \cite{jin2016sso, luo2018improving}.

\textbf{Intrinsic plasticity.} 
Another biologically plausible way of adaptation is based on Intrinsic Plasticity (IP), which is an unsupervised learning rule used for adapting the intrinsic excitability of the reservoir neurons. 
Here, intrinsic excitability refers to a phenomenon called long-term potentiation, in which brief and high-frequent stimulation tends to produce an increased ability to generate spikes \cite{zhang2003other}.
Early research of the integration of RC and IP mainly focuses on reservoir pre-training and global optimizations \cite{steil2007online, schrauwen2008improving, xue2017reservoir}.
In 2019, an IP with a local search scheme was proposed to improve the flexibility of the IP rule by allowing hyperparameters such as learning rate to be different \cite{wang2019echo}.
\textcolor{black}{
In 2022, \cite{nakada2022information} applied IP learning to sucessfully tune the parameter in MEMS-based RC.
}
As a side note, experiments on intrinsic plasticity have shown that the output distributions of real biological neurons may have different forms in different brain regions among various species \cite{li2011model}. 





\section{Recent Approaches in RC} \label{Recent approaches in RC}

\subsection{Overview}
RC has witnessed a significant development in recent years. 
On one hand, traditional RC models such as ESN and LSM have been improved by many new proposed models. 
Some of these models are built on top of the original ones to achieve better performances, while several new architectures of RC have been proposed to solve problems with increasing difficulties. 
On the other hand, recent studies has demonstrated that the idea of reservoir (i.e., a dynamical system that can generate high-dimensional and non-linear responses) can be implemented by using various materials, such as electronic devices, physical systems, and biological realizations.

In this section, we aim to cover recent approaches in RC from several perspectives up until 2023, including ESN-based RC, LSM-based RC, dynamical systems, and physical RC.
These approaches are highly interdisciplinary and are usually tested in several benchmark tasks. 
To introduce some, benchmark tasks for pattern classification includes spoken digit recognition \cite{verstraeten2005isolated}, waveform \cite{paquot2012optoelectronic} and handwritten digit image recognition \cite{jalalvand2015real}. Besides, the non-linear Autoregressive Moving Average (NARMA) time series \cite{jaeger2002adaptive} was widely used in time series forecasting, while a channel equalization benchmark \cite{jaeger2004harnessing} was introduced to evaluate the RC performance on adaptive filtering and control. In addition, temporal XOR task \cite{bertschinger2004real} and memory capacity task \cite{jaeger2002short} were also commonly used in studies that focus on system approximation and short-term memory.

\subsection{Recent Trends of ESN-based RC}
\textcolor{black}{
ESNs represent one of the foundational RC models. The simplicity of implementation makes them an approachable entry point of RC.
Therefore, improving and extending ESNs is not only a key pursuit within the RC community, but its impacts extend far beyond, proving particularly influential for researchers outside the field, and thus allowing more interdisciplinary research. 
In the following, we review recent trends of ESN-based RC models.
}

\subsubsection{Multiple Reservoirs}
\begin{figure*}[ht]
    \centering
    \includegraphics[width=0.99\textwidth]{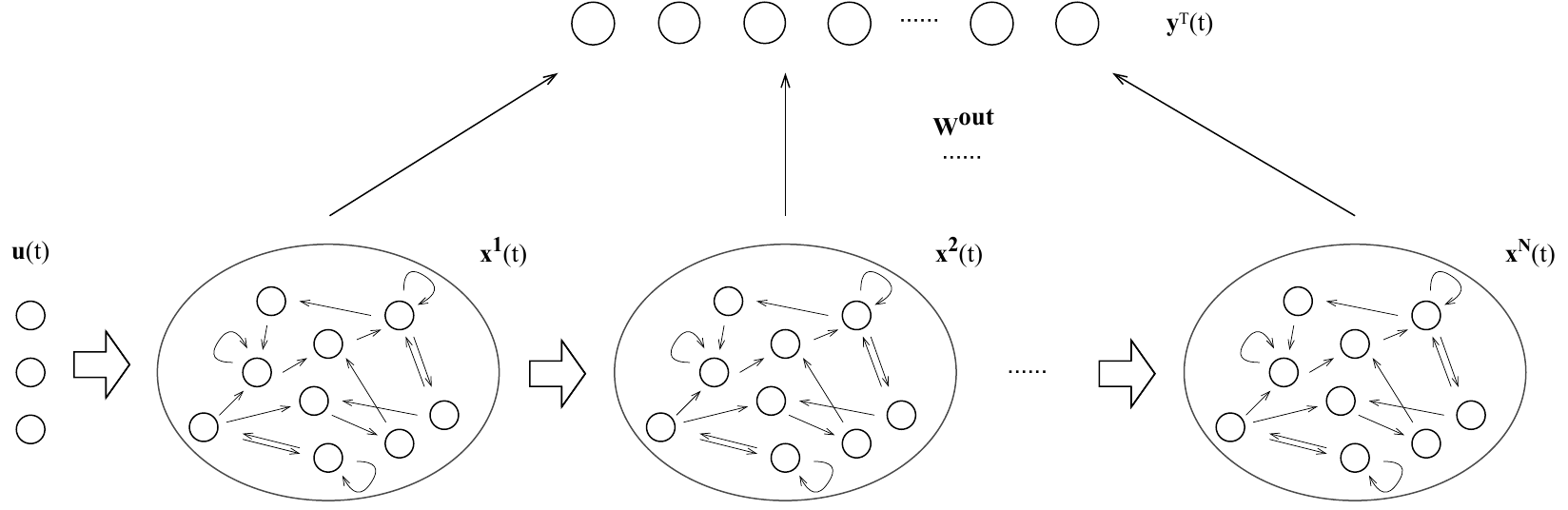}
    \caption{Architecture of DeepESN. For the readout organization for DeepESN, at each time step the reservoir states of all layers are used as input to the output layer. Figure modified from \cite{gallicchio2017deep}.
    }
    \label{fig:DeepESN}
\end{figure*} 

\par
\textbf{Deep ESN.} Apart from the performance saturation problems, there is another limitation in the conventional ESN, i.e., the single large-scale reservoir is poor in simultaneously dealing with different timescales \cite{chitsazan2019wind}. In this concern, some studies started to investigate multiple timescale dynamics of reservoir structure, as it has been found that stacking recurrent networks with different topologies can generate multiple timescales at different layers \cite{gallicchio2017deep}.

In 2016, \cite{gallicchio2017deep} proposed a deep reservoir computing model to achieve hierarchical timescale representation. This model, called deep Echo State Network (DeepESN), stacks multiple reservoirs one on top of the other, as shown in Fig. \ref{fig:DeepESN}. 
\textcolor{black}{Mathematically, consider $N$ the number of reservoir layers. The update equations, extended from Eq. \ref{eq:esn1}-\ref{eq:esn2}, is given by:
}

\textcolor{black}{
\begin{equation}
    \label{eq:deepesn1}
    \mathbf{\Tilde{x}}^{(i)}(n)=tanh(\mathbf{W}^{(i)}_{fwd}\mathbf{x}^{(i-1)}(n)+
        \mathbf{W}^{(i)}_{rec}\mathbf{x}^{(i)}(n-1)),
\end{equation}
}

\textcolor{black}{
\begin{equation}         
\label{eq:deepesn2}
        \mathbf{x}^{(i)}(n) =(1-\alpha^{(i)})\mathbf{x}^{(i)}(n-1) + \\ \alpha^{(i)}\mathbf{\Tilde{x}}^{(i)}(n),
\end{equation}
}

\noindent where 
\textcolor{black}{
$\mathbf{\Tilde{x}}^{(i)}(n)$, $\mathbf{x}^{(i)}(n)$ and $\alpha^{(i)}$ are the update, the reservoir state vector, and the leaking rate at layer $i$, respectively. $\mathbf{W}^{(i)}_{fwd}$ is the weight matrix connecting layers $i$-1 and $i$, and $\mathbf{W}^{(i)}_{rec}$ is the weight matrix of the reservoir at layer $i$, $i=1...N$. The number of input layer is denoted by $i=0$ and $\mathbf{x}^{(0)}(n) = \mathbf{u}(n)$. At each time step $n$, the composition of the states in all the reservoir layers $\mathbf{x}(n)$ is given by: 
}

\textcolor{black}{
\begin{equation}         
\label{eq:deepesn3}
    \mathbf{x}(n) = \langle\mathbf{x}^{(1)}(n), ..., \mathbf{x}^{(N)}(n) \rangle.
\end{equation}
}

Experiments show that the deep RC structure can achieve
(1) multiple timescale representation, where the timescales are ordered along the network's hierarchy;
(2) multiple frequency representation, where progressively higher layers focus on progressively lower frequencies.
Additionally, when there is a perturbation at the input, the effects of this perturbation last longer for higher layers in the stack, and this differentiation is drastically attenuated when input is provided to every layer.
Therefore, having deeper layers while increasing distance from the input is a key architectural factor for obtaining a time-scales separation. 

DeepESN shows potential for designing more efficient RC learning algorithms used for sequential and temporal data processing.
From 2022, A branch of deep ESN is recently studied by the research team of Tanaka et al. modified the deep network architectures were proposed, combining with other techniques such as sequence resampling \cite{li2022multi} in 2022 and Hodrick–Prescott filter \cite{li2023multi} in 2023. These model are claimed to have high prediction performance in time-series prediction tasks with relatively low training cost.

\textbf{Deep Fuzzy ESN.}
In 2019, \cite{zhang2019deep} proposed a novel deep ESN model with fuzzy tuning called Deep Fuzzy ESN (DFESN). Here, two reservoirs are stacked, where the first reservoir is applied for feature extraction and dimensional reduction, and the second one is used for feature reinforcement based on fuzzy clustering. 
In other words, the output of the previous reservoir was extracted as features for the next reservoir input, followed by a feature reinforcing process performed by fuzzy clustering for classification enhancement. 
In DFESN, back propagation is no longer necessary, since the feature reinforcement process can be considered as a layer-wise fuzzy tuning that replaces the back propagation algorithm with lower computational costs.
As claimed by the author, input samples are clustered more easily, thus improving the final classification performance. 

\subsubsection{ESN with Evolutionary Algorithms}
\textbf{Multi-layered echo state network autoencoder.}
Autoencoder (AE) is a type of common feed-forward network for dimensionality reduction and feature detection, in which non-linear transformations are performed in each hidden layer to regenerate a new effective data representation from the originals.
This technique was introduced to RC area by \cite{chouikhi2019bi} as the first recurrent and non-gradient descent-based AE in the literature. In \cite{chouikhi2019bi} an autoencoder was implemented by using multilayered ESN, with a bi-level evolutionary algorithms for optimizing the network architecture and weights. 
Particularly, PSO was applied for the bi-level optimization, where the first level is the architecture determination and the second one is the weights optimization. Classification results on various benchmarks showed that the performance of the evolved model is improved compared with the conventional ESN as well as other CNN or SVM based models.

\textbf{Competitive swarm optimizer.}
Pre-training an ESN using PSO introduces some extra hyperparameters, which are usually determined empirically.
Furthermore, when dealing with high-dimensional optimization tasks, PSO is likely to experience stagnation or premature convergence \cite{long2019evolving}.
To address this, \cite{long2019evolving} designed a deep ESN model with a competitive swarm optimizer (CSO) and used it for fault diagnosis—a precise classification task. 
CSO avoids the problem of optimizing too many parameters at once in PSO with its powerful particle update rule: the particles are updated by evaluating a pre-defined fitness function, and the winner particle will go straight into the next iteration.
For the implementation, CSO is combined with a local search technique to further optimize the deep ESN structure. 
The work shows that deep reservoir networks based on evolutionary algorithms are suitable not only for time series prediction but can also be used to deal with classification problems with adequate results. 

\subsubsection{Other Types of ESNs}
\textbf{Non-linear functions readout.}
As mentioned earlier, single reservoir ESN may not be able to create rich enough non-linear dynamics. 
\cite{chitsazan2019wind} proposed a new method called Non-linear ESN based on non-linear functions and successfully decreased the internal states of the network while increasing dynamic complexity, thus reducing the computational load. 
\textcolor{black}{
Specifically, recall that $\mathbf{x}(n)$ is the internal reservoir state vector ($\mathbf{x}$ for short),} while in this method, it is replaced by a non-linear function:

\textcolor{black}{
\begin{equation}
    \label{eq:nesn}
    \mathbf{x}_{NESN}=\mathbf{f}(\mathbf{x})=
    a_0 + a_1 \mathbf{x} + a_2 \mathbf{x}^2+\cdots+ a_n\mathbf{x}^n.
\end{equation}
}

\noindent where 
\textcolor{black}{$a_n$ being the constant.
With this modification, }non-linear complexity and the learning capability are increased, which results in a higher accuracy in time series forecasting. Moreover, as this method remains simple structure design, it does not require extensive training, parameter tuning or complex optimization process.

\textbf{Small-world topology.}
Small-world (SW) network was first proposed by \cite{watts1998collective}. 
For a regular topological network, each node is usually connected to its neighboring nodes. 
For the connection to other randomly chosen nodes (not adjacent), we denote the connection probability as $p$, where $p = 0$ remain regular topology, $p = 1$ remain random topology and $p \approx 0.1$ as the SW topology (see Fig. \ref{fig:TRNN and SW topology}).

To further investigate the echo state property and learning performance of ESNs, \cite{kawai2019small} presented an SW based ESN (SW-ESN). In this study, the input and output nodes are segregated, and the reservoir remains as an SW topology; that is, neurons connected to the input are different from neurons connected to the output. 
Experiments showed that the SW topology enables the input to flow to the output nodes, and the cluster organizations of the topology guarantee a larger range of echo state property, thus improving the robustness and learning performance of the ESN.

\subsection{Recent Trends of LSM-based RC}
\textbf{Spike-timing-dependent plasticity (STDP).}
STDP is a local unsupervised self-organizing learning rule based on Hebbian learning \cite{jin2016sso}. The main idea of STDP is that if the firing neuron $A$ tends to induce/inhibit spikes from another neuron $B$, the synaptic connection $\mathbf{w}$ from $A$ to $B$ is likely to be potentiated/depressed \cite{zhang2015digital}. In other words, the synaptic connection $\mathbf{w}$ from $A$ to $B$ is potentiated if a causal order (i.e., the presynaptic neuron fires before the postsynaptic neuron) is observed, or depressed otherwise. 
As a spiking neural network, LSM was shown able to be trained by this adopted learning rule in an online learning manner, and therefore significantly reducing memory storage load and computational cost, as well as making LSM becomes more hardware-friendly for physical implementations \cite{jin2016sso, luo2018improving}.
Specially, a recent hands-on implementation of LSM based on \cite{kaiser2017scaling} is realized using NEST simulator \cite{gewaltig2007NEST}.

\textbf{Evolutionary algorithms.}
Similarly to the small world topology, the percentage connectivity indicates the connection probability between neurons within the liquid. Finding a proper percentage of connectivity is then an important factor for improving the accuracy of LSM. Too high/low connectivity will harm the performance, which also suggests that there is an optimal connectivity for a given task.
Particularly, \cite{reynolds2019intelligent} proposed an evolutionary algorithm to optimize the number of neurons and percentage connectivity on a single liquid. Meanwhile, \cite{zhou2019evolutionary} used a covariance matrix adaptation evolution strategy to optimize three parameters, i.e., percentage connectivity, weight distribution and membrane time constant in one liquid. However, \cite{tian2021neural} pointed out that the above algorithms ``only perform parameter optimization in a single liquid and do not optimize the architectures of LSM,” and proposed a Neural Architecture Search (NAS) based framework to optimize both architecture and parameters of LSM model. Furthermore, the presented framework introduced a three-steps search for LSM, where the first step is architecture optimization, the second step is the search for the number of neurons and the final step is parameters optimization. Experimental results show that the proposed model achieves comparable accuracy on classification tasks of the three datasets (i.e., MNIST, Noisy MNIST and FSDD). 


\subsection{Dynamical Systems}
\textcolor{black}{
The key essence of RC lies in its approach to use large, fixed random networks, i.e., the \textit{reservoirs}, exhibiting a rich set of dynamical behaviors. 
These reservoirs provide high-dimensionality and memory in which input data can be transformed and stored, making it easier to model complex temporal patterns and perform machine learning tasks. On the other hand, dynamical systems, characterized by their temporal evolution and behavior, offer crucial insights into the working of these reservoirs, shaping how we understand and optimize them. 
In the following subsections, we show how RC provides a practical framework for studying dynamical systems, while the theories underlying dynamical systems give a solid mathematical foundation to the operation of RC. 
}

\begin{figure*}[ht]
    \centering
    \includegraphics[width=1\textwidth]{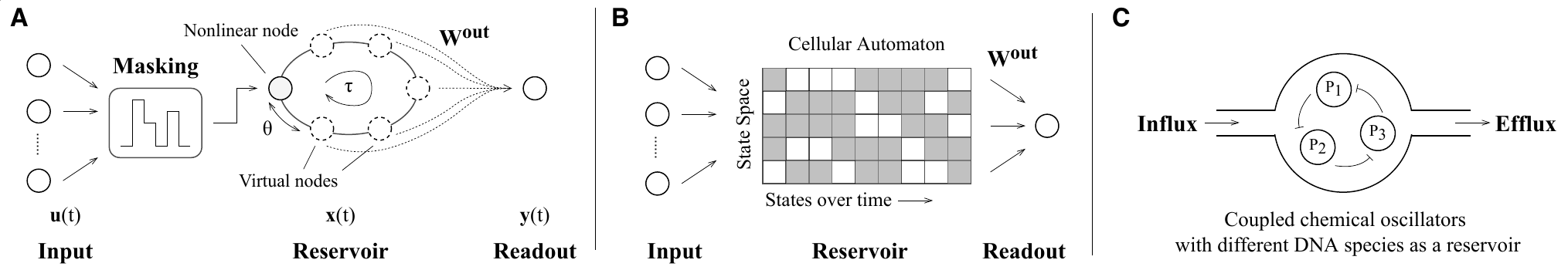}
    \caption{RC with dynamical systems.
    (A) Single-node time-delayed feedback reservoir \cite{appeltant2011information}. 
    (B) Cellular automaton \cite{yilmaz2015machine}.
    (C) DNA oscillator \cite{goudarzi2013dna}.
    }
    \label{fig:dynamical systems}
\end{figure*}

\subsubsection{Single-node Time-delayed Feedback Reservoir}
\label{sec:sntdfbrc}
\textbf{Definition.} Classical RC models process a low-dimensional temporal input through a high-dimensional reservoir state space. This high-dimensional state space is achieved by creating many randomly connected artificial (ESN) or spiking (LSM) neurons as a reservoir, so as to receive input data coupled into the reservoir with weight (synaptic) connections. 
It turns out that RC can be implemented by using only a single hardware node. In 2011, the concept of using a single dynamical node as a reservoir to generate a high-dimensional space, was introduced in \cite{appeltant2011information}. 
In contrast to network-based reservoirs consisting of many neurons such as ESN and LSM, single node RC shows great simplicity especially for physical implementation, 
Theoretically, the proposed delay system refers to non-linear delayed feedback system, which is a type of dynamical system described by delay differential equations given by \cite{lepri1994high}:
\begin{equation}
    \frac{dx(t)}{dx}=F(t,x(t),x(t-\tau)),
\end{equation}

\noindent where $t$ and $x$ are continuous time and state variables, respectively. $F$ is the function of this system with $\tau>0$ the delay factor. This system is usually employed by using electronic circuits with a feedback loop (see Fig. \ref{fig:dynamical systems}A). 
\textcolor{black}{
The workflow is as follows:
\begin{itemize}
    \item A low-dimensional input signal (e.g., one-dimensional temporal signal) is first processed using a time-multiplexing masking function, and modulates the state of the node.
    \item The single node samples the pre-processed input states and holds them for a delay period of $\tau$. 
    \item Meanwhile, $N$ virtual nodes are set that equally divide $\tau$ with the time interval of $\theta=\tau/N$, forming a delay line.
    \item When the signal reaches the end of the delay line, it is fed back into the node, influencing the node's future states.
    \item The current state of the node and the states stored in the delay line are then fed to the readout layer with trainable weighted connections.
\end{itemize}
}

\textbf{Extensions.} The single-node delayed feedback RC was experimentally investigated on spoken digit recognition task and NARMA time series prediction task \cite{verstraeten2005isolated, jaeger2002short}. 
Some variations of delayed feedback structure were proposed, such as
(1) Different ensembles of delay-based RC with several delayed neurons by \cite{ortin2017reservoir}. 
and (2) two circular connected time-delayed based reservoirs with a longer delay line by \cite{brunner2018tutorial}.

\subsubsection{Cellular Automaton}
Another type of dynamical system that can be used as an RC is the cellular automaton (CA). 
CA is a collection of a cell-grid of specified shape that evolves (interacts with its neighbors and changes its state) through a number of discrete time steps \cite{wolfram2018cellular}. CA's new state is determined by a pre-defined set of update rules and neighboring cells, resulting in a rich dynamic (see Fig. \ref{fig:dynamical systems}B).

The concept of CA system was further extended to RC. A series of work on building CA-based RC was made by \cite{yilmaz2014reservoir, yilmaz2015machine}, where an evolution rule was introduced to create a space-time volume in the automaton state space (i.e., the reservoir). 
\textcolor{black}{
The proposed CA system was reported to be suitable for combining other types of discrete dynamical systems such as 
Boolean logic and symbolic processing \cite{snyder2013computational}.
}
Recent studies on CA-based RC also include modifications and extensions of network architecture \cite{nichele2017reservoir, nichele2017deep}; 
as well as improvements of evolution rules in CA \cite{mcdonald2017reservoir}.
\textcolor{black}{More recently, \cite{uragami2022universal} explored the advantages of critical spacetime patterns generated by elementary cellular automata (ECAs) in reservoir computing, specifically focusing on the distractor's length in time series data and proposing asynchronously tuned ECAs (AT-ECAs) to generate universally critical spacetime patterns.}

\subsubsection{Coupled Oscillators}
\textbf{Definition.} 
RC models have been successfully applied in
research areas such as machinery, chemistry, biology and physical systems (see section \ref{Recent Applications of Reservoir Computing}). Among these implementations, many RCs are built using coupled oscillators. A general representation of the dynamics of coupled oscillators is given by an ordinary differential equation:

\textcolor{black}{
\begin{equation}
    \frac{dx_{i}(t)}{dt} = F(x_{i}(t))+
    G(x_{1}(t),...,x_{N}(t)),
\end{equation}
}

\noindent where $i=1,...,N$ is the index of a total of $N$ coupled oscillators, $x_{i}(t)$ is the state of $i^{th}$ the oscillator at time $t$, $F$ and $G$ are an isolated function and a coupling function. The rich dynamics are provided by each oscillator and the interactions between them \cite{tanaka2019recent}.
\textcolor{black}{
In the following, we aim to review studies that utilize coupled oscillators as a building block of RC. Additionally, we will discuss oscillation mechanism of brain in Section \ref{RC with brain mechanisms and cognitive science}.
}

\textbf{Mechanical oscillators.} 
The first category of RCs using coupled oscillators is based on mechanical oscillators. 
\cite{coulombe2017computing} built a network with \textit{anharmonic} (i.e., non-linear) oscillators, where the components include masses coupled linear or non-linear springs, making the system power-efficient to solve a bit-stream computation task and a spoken words classification task.

\textbf{DNA oscillators.} 
In the field of molecular computing, a deoxyribonucleic oscillator (DNA) reservoir was first proposed in \cite{goudarzi2013dna}. 
This RC consists of coupled deoxyribozyme based oscillators. 
Specifically, a microfluidic reaction \textit{chamber} was used to construct a reservoir, since different DNA species can interact (see Fig. \ref{fig:dynamical systems}C).
\textcolor{black}{
Here, the microfluidic reaction chamber is a specific kind of chamber used to carry out chemical or biological reactions under well-controlled conditions.
}
Rich transient dynamics were then generated in the reservoir, where the reservoir state consists of the time-varying concentration of various species inside the chamber. 
A signal-tracking task was then performed by using a reservoir with three DNA species that exhibits oscillatory behavior.
Recent developments of DNA oscillators include a random chemical RC model \cite{nguyen2020reservoir}, where the random chemical circuits (i.e., DNA strand displacement) provide complex non-linear dynamics, making them suitable for RC implementation. 
A novel RC using DNA oscillators was reported in \cite{liu2022reservoir} which solves the problem of the lack of readout layer in the previous work \cite{goudarzi2013dna}, and was then demonstrated for the handwritten digit recognition and a second-order non-linear prediction.

\textbf{Chemical reaction networks (CRNs).}
Related to the DNA oscillators, the chemical reaction networks also show the capabilities to RC implementations.
One of the initial studies on CRN-based RC model was presented in a presentation \cite{nguyen2018biochemical}, from which the reservoir dynamic is given by a set of ordinary differential equations (ODEs), while the readout layer is to learn the Hamming distance between input bit-streams. 
In 2022, the author further proposed a chemical RC for single stranded DNA (ssDNA) analysis \cite{nguyen2022sers}.
Additionally, \cite{yahiro2018reservoir} used a modular framework to implement a RC model. 
The main advantage of this work, compared with the previous DNA oscillators \cite{goudarzi2013dna}, is that molecular computing allows changing the size of CRNs on-the-fly.
Another new chemical RC architecture was proposed by \cite{kan2021physical}, where the reservoir was implemented through electrochemical reactions.
Also, it is reported that the Polyoxometalate molecule (POM) in this chemical RC increases the diversity of the response current and thus improves their abilities to predict periodic signals. 
POM-based RC was further integrated with the so-called \textit{single-walled carbon nanotubes} as a random dense network \cite{akai2022performance}. Adequate results were obtained in tasks including waveform reconstruction, non-linear autoregressive modelling and memory capacity testing.

\textbf{Other RCs with oscillators.}
It is reported that oscillatory behavior can be restricted to the phase domain \cite{nakao2016phase}. This makes it possible to apply phase oscillators that exhibit rich dynamics to RC \cite{yamane2015wave}.
A RC using two coupled relaxation oscillators built on $VO_2$ switches was reported \cite{velichko2020reservoir}, where the oscillators show high order synchronization that allows simulating the XOR operation.
Besides, RC can also be implemented by using spin-torque nano-oscillators in neuromorphic computing \cite{riou2017neuromorphic}. See Section \ref{Physical_RC} for detail of spintronic RC.


\begin{table*}[]
\centering
\renewcommand{\arraystretch}{1.0}
\caption{Types of physical implementations of RC: components and applications.}
\resizebox{0.99\textwidth}{!}{%
\begin{tabular}{llll}
\hline \hline
\textbf{RC Type} &
  \textbf{Name} &
  \textbf{Components / Methods} &
  \textbf{Benchmark tasks / Applications}
     \\ \hline \hline
\multirow{3}{*}{\textbf{Electronic RC}} &
  Analog Circuits &
  \begin{tabular}[c]{@{}l@{}}1. Various electronic elements \\ with digital hardware.\\ 2. Spiking circuit implementations.\end{tabular} &
  \begin{tabular}[c]{@{}l@{}}Spoken digit recognition \cite{appeltant2014constructing}. \\ Memory capacity estimation \cite{appeltant2014constructing}.\\ Time-series prediction \cite{li2018deep}. \\ ECG signal processing \cite{li2018deep}. \\ Non-temporal non-linear task \cite{jensen2017reservoir}. \\ Efficient spiking implementation \cite{zhao2016novel, li2017analog}. \end{tabular} 
  \\ \cline{2-4} 
 &
  FPGAs &
  \begin{tabular}[c]{@{}l@{}}1. FPGAs board with stochastic logic.\\ 2. Recurrent SNN on FPGA.\\ 3. Parallel neuromorphic hardwares.\end{tabular} &
  \begin{tabular}[c]{@{}l@{}}Channel equalization problems \cite{antonik2015fpga}. \\ Image and isolated digit recognition \cite{schrauwen2008compact, wang2016liquid}. \\ Short input and waveform- \\ patterns classification \cite{haynes2015reservoir, alomar2015digital}. \end{tabular}
  \\ \cline{2-4} 
 &
  Memristor &
  \begin{tabular}[c]{@{}l@{}}Neuromemristive components:\\ 1. Double crossbar arrays.\\ Memristive devices without neurons:\\ 1. Random memristor networks.\\ 2. Memristor with volatility.\\ 3. Memcapacitors. \\ 4. Atomic switch networks.\end{tabular} &
  \begin{tabular}[c]{@{}l@{}}Time-series prediction \cite{yang2016investigations}. \\ Waveform pattern generation/classification \cite{kulkarni2012memristor}. \\ Associative memory task \cite{kulkarni2012memristor}. \\ Image recognition \cite{du2017reservoir}.\end{tabular} 
  \\ \hline
\multirow{2}{*}{\textbf{Photonic RC}} &
  \begin{tabular}[c]{@{}l@{}}Spatially distributed \\ Optical nodes\end{tabular} &
  \begin{tabular}[c]{@{}l@{}}1. Semiconductor optical amplifiers \\ (SOAs) with digital masking. \\ 2. Photonic crystal platform.\\ 3. Nodes with free-space optics principles.\end{tabular} &
  \begin{tabular}[c]{@{}l@{}} Optical packet header identification \cite{vandoorne2014experimental}. \\ Spoken digit classification \cite{katumba2018low}.\\ Logical function prediction \cite{dong2018scaling}. \\ Waveform prediction \cite{fiers2014nanophotonic}. \\ Memory capacity estimation \cite{laporte2018numerical}.\end{tabular} \\ \cline{2-4} 
 &
  \begin{tabular}[c]{@{}l@{}}Time-delayed \\ feedback loop\end{tabular} &
  \begin{tabular}[c]{@{}l@{}}1. Opto-electronic feedback loop. \\ 2. All-optical reservoir. \\ 3. Coherently driven passive cavity.\end{tabular} &
  \begin{tabular}[c]{@{}l@{}}Non-linear channel equalization \cite{vinckier2015high}. \\ Spoken digit recognition \cite{vinckier2015high}.\end{tabular} \\ \hline
\multirow{4}{*}{\textbf{Spintronic RC}} &
  Spin-torque oscillators &
  Magnetic tunnel junction. &
  \begin{tabular}[c]{@{}l@{}}Short-term memory task \cite{taniguchi2021reservoir}. \\ Macro-magnetic simulation \cite{furuta2018macromagnetic}.\end{tabular} \\ \cline{2-4} 
 &
  Spin wave &
  \begin{tabular}[c]{@{}l@{}}Thin Yttrium iron garnet film between \\ a magneto-electric coupling layer \\ and a conductive substrate.\end{tabular} &
  Temporal XOR problems \cite{nakane2021spin}. 
   \\ \cline{2-4} 
&
  Magnetic skyrmions &
  \begin{tabular}[c]{@{}l@{}}1. Nanoscale magnetic vortex. \\ 2. Skyrmion fabrics.\\ 3. Magnetic skyrmion memristor.\end{tabular} &
  \begin{tabular}[c]{@{}l@{}}Image classification task \cite{jiang2019physical}. \\ Handwritten digit recognition \cite{jiang2019physical}.\end{tabular}
   \\ \cline{2-4} 
 &
  \begin{tabular}[c]{@{}l@{}}Dipole-Coupled\\ Nanomagnets\end{tabular} &
  \begin{tabular}[c]{@{}l@{}}1. Magnetic nonodots array. \\ 2. Magnetic random access memory \\ (MRAM) technology.\end{tabular} &
    \begin{tabular}[c]{@{}l@{}} NARMA10 task \cite{nomura2021reservoir}. \\ large-scale RC implementation \cite{nomura2021reservoir}. \end{tabular}   \\ \hline
\multirow{3}{*}{\textbf{Mechanical RC}} &
  \begin{tabular}[c]{@{}l@{}}Mass-spring-damper \\ systems\end{tabular} &
  \begin{tabular}[c]{@{}l@{}}1. Mass-spring-damper simulating\\  softbodied robots.\\ 2. Soft robotic arm (octopus). \\ 3. Pneumatically driven robotic arm.\end{tabular} &
  \begin{tabular}[c]{@{}l@{}}Active shape discrimination \cite{johnson2014active}. \\ Learning to emulate timers- \\ delays and parity \cite{nakajima2014exploiting, nakajima2018exploiting}.\\ Robot crawling \cite{bhovad2021physical}.\\ Foraging learning task \cite{yada2021physical}.\end{tabular} \\ \cline{2-4} 
 &
  Sensors &
  \begin{tabular}[c]{@{}l@{}}State Weaving Environment Echo \\ Tracker (SWEET) sensing.\end{tabular} &
  Ion concentration analysis \cite{konkoli2018developing}. 
   \\ \cline{2-4} 
 &
  \begin{tabular}[c]{@{}l@{}}Tensegrity robots and \\ Central pattern generator\end{tabular} &
  \begin{tabular}[c]{@{}l@{}}1. Tensegrity-based structure with tensile \\ elements and compressive elements. \\ 2. Spiking-based reservoir acting as CPG.\end{tabular} &
  Locomotion and sensing tasks \cite{caluwaerts2011body}. \\ \hline
\textbf{Quantum RC} &
  \begin{tabular}[c]{@{}l@{}}Quantum circuits / \\ computers\end{tabular} &
  \begin{tabular}[c]{@{}l@{}}1. Analog quantum dynamics under \\ a time-dependent Hamiltonian. \\ 2. Nuclear magnetic resonance (NMR).\\ 3. Quantum circuits with quantum gates.\end{tabular} & 
  \begin{tabular}[c]{@{}l@{}} Quantum-chaotic system- \\ implementation \cite{ghosh2021realising, govia2021quantum, martinez2021dynamical}. \end{tabular}
 \\ \hline
\end{tabular}%
}

\end{table*}



\subsection{Physical RC}
\label{Physical_RC}
Recall that the key feature of RC models is to transform sequential/temporal inputs into a high-dimensional non-linear dynamical space (i.e., the reservoir). If the reservoir provides rich enough dynamics, the desired output can be read out by using simple learning methods such as linear regression (see section \ref{Training a RC model} for details). Therefore, in principle, any kind of “non-linear, high-dimensional dynamical systems which satisfies some conditions”, has the potential to be a reservoir.

In particular, RC has become popular in a wide range of research fields focusing on hardware design; that is, recent trend of RC implementation have shifted to many domains of physical reservoir computing (PRC) such as optical systems, neuromorphic devices, chemical reactions, quantum computing, to name a few. 

Several reviews have tried to organize this highly interdisciplinary topic of PRC. 
A comprehensive overview of recent PRC implementations was reported in \cite{tanaka2019recent}, while 
\cite{van2017advances} focuses on the recent advances in photonic RC. 
A book series with detailed and special issues on designing PRC was published \cite{nakajima2021reservoir}. 
In the following sections, we briefly outline the recently proposed PRC models for completeness. Readers may refer to the articles above for a more systematic and theoretical understanding of PRC.


\subsubsection{Electronic RC}
\begin{figure*}[ht]
    \centering
    \includegraphics[width=0.99\textwidth]{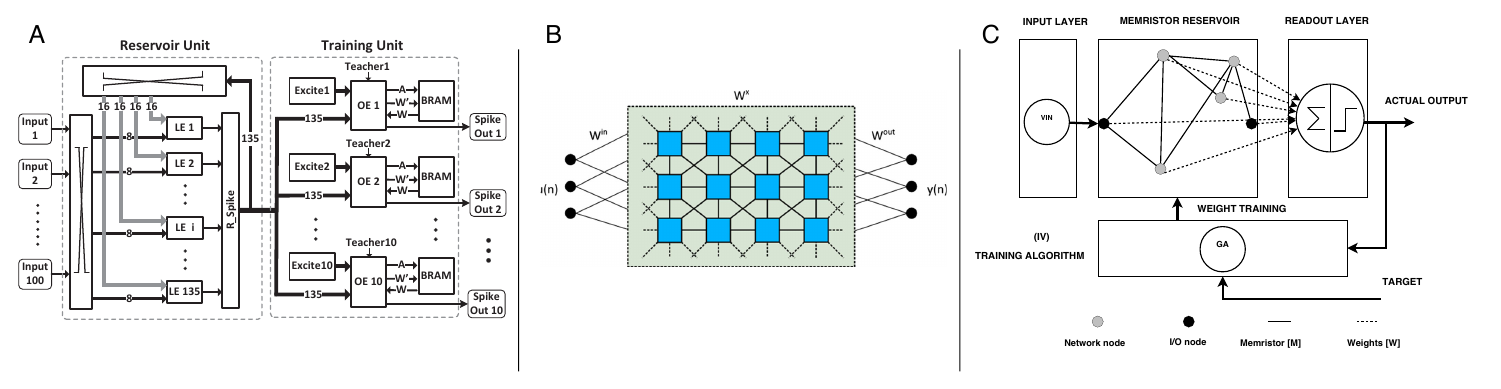}
    \caption{Electronic RC.
    (A) Hardware LSM architecture using FPGAs. BRAM refers to block RAM, where $W$ and $W'$ are the old and updated synaptic weights, respectively \cite{wang2016liquid}.
    (B) RC implemented by memristors with local connections \cite{yang2016investigations}.
    (C) Memristor-based reservoir computing using GAs for training \cite{kulkarni2012memristor}.
    }
    \label{fig:Electronic RC}
\end{figure*}

\textbf{Analog circuits.} 
Various electronic circuits are potential components for hardware RC. Previously, we discuss the single-node time-delayed feedback RC that was first proposed by \cite{appeltant2011information, appeltant2014constructing}. 
In fact, this type of RC can be formed by electronic circuits with other digital hardware elements \cite{soriano2014delay, soriano2015minimal}.  Additionally, a single-node RC implemented by using spiking circuits was also reported \cite{zhao2016novel, li2017analog}. 
The advantages of analog circuit based RC include (1) less hardware requirements and (2) power efficiency in spiking implementations.
Therefore, RCs based on analog circuits were successfully applied to tasks including
(1) spoken digit recognition and memory capacity estimation  \cite{appeltant2014constructing};
(2) time series prediction and ECG signal processing \cite{li2018deep};
and (3) non-temporal non-linear tasks \cite{jensen2017reservoir}.

\textbf{Field-programmable gate array (FPGA).}
FPGA board, as a hardware friendly element, has been proven to be suitable for RC implementations.
An early attempt at the combination of RC and FPGA board was reported by \cite{verstraeten2005reservoir}. 
In 2014, an FPGA board with stochastic logic was used to implement RC for non-linear time series prediction task \cite{alomar2014low}. 
The offline learning in the work above was further modified to an online learning scheme \cite{antonik2015fpga}, where the units in the reservoir exhibit sigmoid activation and learn with a gradient descent algorithm.
In terms of spiking implementation, the conventional LSM models were successfully built on FPGAs in an early research \cite{schrauwen2008compact} (see Fig. \ref{fig:Electronic RC}A), as well as some models featured with parallel processing \cite{wang2016liquid} and STDP learning rule \cite{liu2018online}.
In fact, FPGA-based RC have shown 
(1) better re-configurability;
(2) much faster processing speed;
(3) less energy costs compared to general CPUs;
and (4) more biologically plausible (SNN-based LSM).
Several benchmark tests were taken, indicating FPGAs based RC can be applied to
(1) channel equalization problems \cite{antonik2015fpga};
(2) image and isolated digit recognition \cite{schrauwen2008compact, wang2016liquid};
and (3) short input and waveform patterns classification \cite{haynes2015reservoir, alomar2015digital}.

\textbf{Memristive RC.}
\textcolor{black}{A memristive device, or memristor, is a type of passive circuit element that maintains a relationship between the time integrals of current and voltage across a device.}
Some studies of RC focus on using memristive elements.
Here, the property of memristive elements is different from other circuit ones, since they vary the resistance depending on the current flow at different times. 
There are two main types of memristive components suitable for RC. 
The first type is based on neuromemristive circuits.
Specifically, memristors are used to model the synaptic plasticity between neurons.
\cite{yang2016investigations} successfully built an ESN based on memristor (see Fig. \ref{fig:Electronic RC}B), yet the performance was worse than the conventional ESN in terms of time-series prediction task.
Other studies include using double crossbar arrays as reservoir in ESNs \cite{hassan2017hardware} and LSMs \cite{soures2017robustness}.

Another branch of studies has shown that memristive devices without neurons can also generate rich non-linear dynamics for RC implementations.
The first attempt was made by \cite{kulkarni2012memristor} for a wave pattern classification task, where a memritstive topology was applied as a reservoir (Fig. \ref{fig:Electronic RC}C). In 2017, 
\textcolor{black}{another} RC model using dynamic memristors for hard digits recognition was reported \cite{du2017reservoir}.
Other memristive networks were also explored to have potential of constructing RC, such as 
(1) random memristor networks \cite{burger2015computational};
(2) memristor with volatility \cite{carbajal2015memristor};
(3) memcapacitors \cite{tran2017memcapacitive} and its hierarchy extension \cite{tran2019hierarchical};
and (4) atomic switch networks \cite{stieg2012emergent}.



\subsubsection{Photonic RC}
Optical computing is another paradigm suitable for RC implementations, where the complex non-linear and high-dimensional dynamics can be achieved in the intensity and phase of the optical field. 
A wide range of studies aim to uncover this specific area of RC. 
In principle, there are two main directions of photonic RC implementations: 
(1) spatially distributed optical nodes, and 
(2) time-delayed based photonic RC.
For more detailed investigation and discussion, please refer to two comprehensive reviews in \cite{van2017advances, tanaka2019recent}.

\textbf{Spatially distributed optical nodes.} 
It has been realized that the fixed and randomly connected topologies in the conventional RC models (e.g., ESN and LSM) can be implemented by spatially extended photonic networks using spatially distributed optical nodes.
Perhaps the first optical RC was proposed and subsequently developed in  \cite{vandoorne2008toward, vandoorne2014experimental}. 
Here, an on-chip network of semiconductor optical amplifiers (SOAs) was constructed to efficiently compute the \textit{tanh} function in the reservoir.
Later, a digital masking approach was proposed to overcome the short time delay and high operation rate in the previous photonic RC \cite{schneider2016using}.
From 2014 on, several techniques based on optical nodes were reported, including
(1) photonic crystal platform \cite{fiers2014nanophotonic, laporte2018numerical};
and (2) nodes with free-space optics principles \cite{mesaritakis2015high}.
In fact, optical nodes RC are not only operating with lower power consumption, but more importantly, they are extremely fast in computation.
In terms of benchmark tasks and applications, RC with optical nodes were simulated and applied to 
(1) optical packet header identificationand spoken digit classification \cite{vandoorne2014experimental, katumba2018low};
(2) logical function prediction \cite{dong2018scaling};
(3) waveform prediction task \cite{fiers2014nanophotonic};
and (4) memory capacity task \cite{laporte2018numerical}.

\begin{figure*}[ht]
    \centering
    \includegraphics[width=0.97\textwidth]{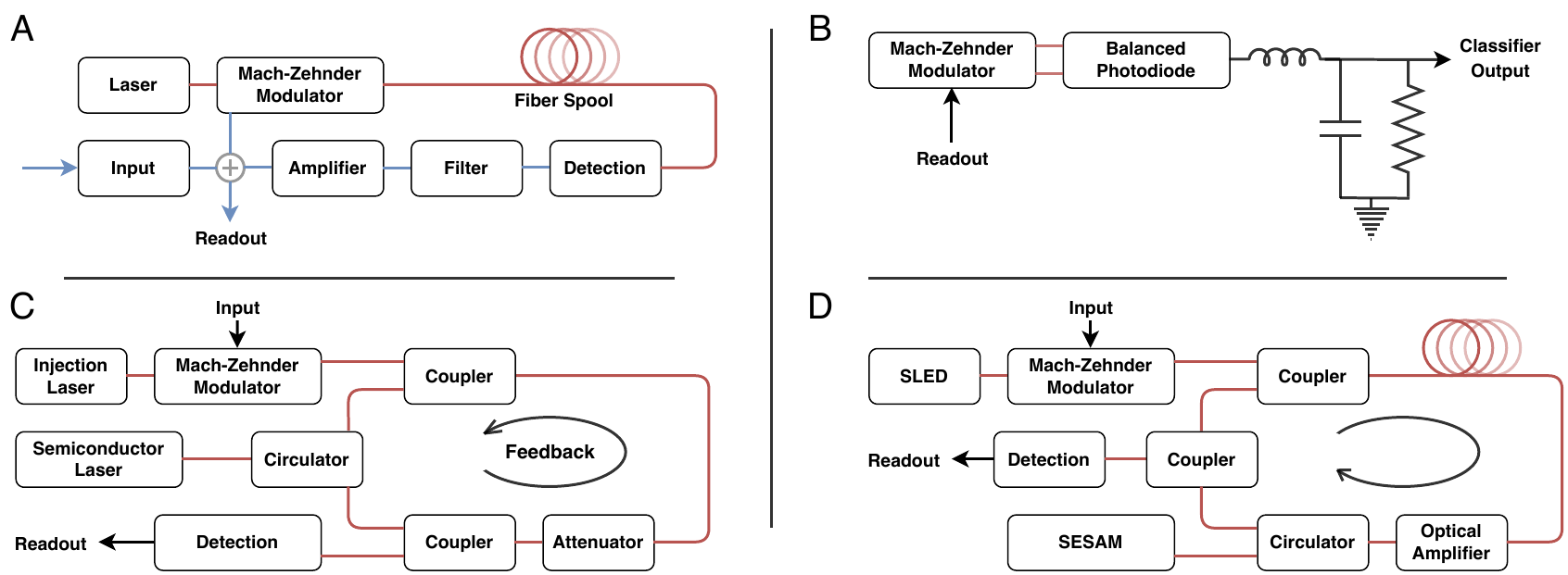}
    \caption{RC implemented with electronic and optical elements \cite{van2017advances}. 
    (A) Opto-electronic RC system. The optical (electronic) path is depicted in red (blue) color.
    (B) Readout layer used in photonic RC by \cite{duport2012all}.
    (C) All-optical RC based on a semiconductor laser subject to delayed optical feedback.
    (D) All-optical RC based on passive devices.
    }
    \label{fig:phtonic RC}
\end{figure*}

\textbf{Time-delayed feedback RC.} 
In section \ref{sec:sntdfbrc}, we review RC with single-node time-delayed feedback loop. After the first electronic time-delayed reservoir proposed by \cite{appeltant2011information}, optical devices were quickly used to implement such systems \cite{larger2012photonic, paquot2012optoelectronic}. 
In specific, these pioneering systems applied opto-electronic feedback loops, where the optical parts with long fiber provide non-linearity and time-delay while the electronic parts play the role of input processing and output extraction \cite{van2017advances}.
Later, the electronic parts of the opto-electronic based RC were replaced by active optical devices (i.e., SOAs or fiber coupler) \cite{duport2012all}, forming an all-optical delayed based RC.
Moreover, it is reported that a significant improvement of high-speed, low-consumption can be achieved by using passive devices (see Fig. \ref{fig:phtonic RC}D). An example is a RC with a coherently driven passive cavity proposed in \cite{vinckier2015high},
in which a simple linear fiber cavity was used as a reservoir to solve tasks such as non-linear channel equalization and spoken digit recognition with remarkable performance.

\subsubsection{Spintronic RC}
\textcolor{black}{Spintronics, or spin electronics, is a branch of physics (particularly condensed matter physics) and nanotechnology that uses the intrinsic spin of the electron and its associated magnetic moment, in addition to its fundamental electronic charge, in solid-state devices.}
It is reported independently that several spintronic elements can be candidates for RC implementations, including but not limited to 
spin-torque oscillators, 
spin waves,
magnetic skyrmion;
and Dipole-coupled nano-magnets.
Spintronic RC were reviewed in \cite{tanaka2019recent}, as well as in \cite{taniguchi2021reservoir} specially for spin-torque oscillators.

\textbf{Spin-torque oscillators.}
The first experiments of building a spintronic RC can be found in \cite{torrejon2017neuromorphic, riou2021reservoir}, where the spin-torque oscillators were used to provide non-linearity. 
Here, the so-called magnetic tunnel junction is the key component and is used as a reservoir.
The advantage of spin-torque oscillators is that the whole neural network can be emulated by the fast magnetization dynamics generated by simple components. 
The performances of spin oscillators based RC were evaluated with short-term memory estimation experiments, and were quantified by macro-magnetic simulation \cite{taniguchi2021reservoir, furuta2018macromagnetic}.

\textbf{Spin waves RC.}
Another branch of spintronic RC is using spin waves as a reservoir. 
The first RC based on spin waves was proposed by \cite{nakane2018reservoir}. A bit sequence input was used and a stripe magnetic domain structure is introduced in a continuous magnetic garnet film where spin waves propagate, thus generating spatially distributed rich dynamics.
In 2021, spin waves RC was numerically evaluated in \cite{nakane2021spin}, 
and was further shown to be low power consuming \cite{taniguchi2022spintronic}.
For the applications, spin waves RC in the above studies shows state-of-the-art performances in numerical experiments of temporal XOR problems and memory capacity tasks, once the model is properly tuned.

\textbf{Magnetic skyrmions.} 
Magnetic skyrmions are small swirling topological defects in the magnetization texture, in which the stabilization and dynamics depend strongly on the topological properties of skyrmions \cite{fert2017magnetic}. 
In 2018, the first prototype of RC based on skyrmion fabrics was proposed in \cite{bourianoff2018potential, prychynenko2018magnetic}. Here, the skyrmion fabrics refer to the phases that interpolate between single skyrmions, skyrmion crystals and magnetic domain walls. Owing to their random phase structures, they are claimed to be suitable for RC implementation. 
Another application on RC based on magnetic skyrmion is to implement physical RC based on a single magnetic skyrmion memristor (MSM) for image classification task (i.e., handwritten digit recognition) \cite{jiang2019physical}.

\textbf{Dipole-coupled nanomagnets.}
A novel magnetic nanodots array was reported for a new way of RC implementations \cite{nomura2021reservoir}. The proposed system is a static magnetic system, in which the dynamics are enriched by increasing the number of nanomagnets. 
Specifically, the reservoir is formed by dipole-coupled nanomagnets (nodes). Similarly to the conventional RNN-based RC, some nodes were connected to the input, while all nodes were connected to the output and were then read out by the magnetic random access memory (MRAM) technology.
A good aspect of nanomagnets based RC is that the magnetic interconnections solve the wiring problem of hardware RNN implementations, showing a great potential to build a large-scale RC system.
The system performance was evaluated in the NARMA10 task with adequate results.

\subsubsection{Mechanical RC}
\textbf{Mass-spring-damper systems.} 
Several mechanical RC models have been proposed, including but not limited to soft robots and sensors networks. 
The idea of employing robot's body and its dynamics as a computational resource for RC, is originated from the early works \cite{hauser2011towards, hauser2012role}.
The so-called mass-spring-damper systems are used to replace the conventional neurons in the reservoir (e.g., artificial nodes in ESNs). Specifically, mass-spring-damper systems serve as good models to simulate the biological bodies and soft-bodied robots, where both systems show rich non-linear dynamics that can be used in physical RC implementations. 
Inspired by the pioneering works above, \cite{johnson2014active} built a mass-spring-damper system for active shape discrimination.
From a more biologically plausible perspective, \cite{nakajima2013soft} created a soft robotic arm based on an octopus. 
The work implies that control can partially be outsourced to the physical body and the interaction with the environment without being processed by the brain or a controller.
The authors further made a series of work on the octopus-inspired robotic RC, showing that the implementations can learn to emulate timers, delays and parity \cite{nakajima2014exploiting, nakajima2018exploiting}.
For more applications of mechanical RC, please see Section \ref{Recent Applications of Reservoir Computing}.


\textbf{Sensors.}
RC had been shown potential for processing sensor data (see section \ref{Recent Applications of Reservoir Computing}). Yet some researchers like \cite{konkoli2018developing} argued that those reservoirs focusing on sensing are often exploited in a somewhat passive manner, being a separated post-processing component that receives data from sensors. 
Therefore, \cite{konkoli2018developing} proposed the State Weaving Environment Echo Tracker (SWEET) sensing approach, where the RC is considered the sensing element itself for novel sensing applications such as ion concentration analysis.

\textbf{Tensegrity robots and central pattern generator (CPG).}
For completeness, tensegrity based robots and CPGs are briefly reviewed here. 
\textcolor{black}{
To illustrate, tensegrity refers to a stable structure that consists of tensile elements connected by additional compressive elements \cite{fujita2018environmental}, which
is considered to be as adaptive and resilient as the biological systems. 
Another concept, the central pattern generator (CPG), is a neural network that can produce rhythmic patterned outputs without relying on rhythmic sensory or central inputs \cite{cpg1, cpg2, cpg3, cpg4}. }
On top of the mass-spring systems, \cite{caluwaerts2011body} first developed a RC based on a tensegrity based structure and applied it to locomotion and sensing tasks. 
Tensegrity structure can be further used as computational resources for modelling biological structures like human bodies and cells due to its stability. 
For example, CPG signals were generated by a tensegrity based RC for locomotion 
\cite{fujita2018physical, caluwaerts2014design}.
Another CPG-related RC was reported in \cite{vandesompele2019populations}, where the FORCE learning algorithm was used to train a spiking-based reservoir that acts as a CPG. 
\cite{terajima2021behavioral} further showed that the biologically plausible tensegrity robots are capable of adaptation to environmental changes.

\subsubsection{Quantum RC}
Quantum reservoir computing (QRC) is an intersection of quantum computing and RC. Reviews on this topic can be found in 
\cite{mujal2021opportunities, fujii2021quantum, ghosh2021quantum}.
The platform of QRC was first proposed by \cite{fujii2017harnessing}.
The idea is to use analog quantum dynamics under a time-dependent Hamiltonian, where the parameters are randomly chosen without tuning. 
\textcolor{black}{
Here, the  Hamiltonian is a mathematical operator used to describe the total energy of a quantum system. It plays a central role in the Schrödinger equation, which describes how a quantum state evolves over time. 
}
Several improvements of QRC have been explored, including
(1) boosting computing power \cite{nakajima2019boosting};
(2) enhancing memory capacity \cite{kutvonen2020optimizing};
and (3) using Nuclear Magnetic Resonance (NMR) \cite{negoro2021toward}.
Taking the advantage that any quantum-chaotic system can be used for implementations, 
QRC are investigated in many other studies, including
(1) using quantum circuits with quantum gates \cite{ghosh2021realising};
(2) single non-linear oscillator \cite{govia2021quantum};
and (3) dynamical phase transitions \cite{martinez2021dynamical}.



\subsection{Other RC models}

\textbf{RC with non-linear vector autoregression (NVAR).}
It turns out that RC can be realized as a general, universal approximator of dynamical systems, in which the RNN part contains non-linear activation neurons while the readout layer is a weighted linear sum of the reservoir states.
A novel concept was proposed by \cite{gonon2019reservoir} and \cite{hart2021echo}, that RC with linear activation of neurons followed by a non-linear readout is equivalent to a universal approximator. 
In this case, such a RC would become mathematically identical to a non-linear vector autoregression (NVAR) machine \cite{bollt2021explaining}.
By identifying the limitations of random reservoir and taking inspiration from \cite{bollt2021explaining}, \cite{gauthier2021next} proposed a so-called next generation reservoir computing (NG-RC) model based on NVAR.
The proposed model was built without the requirements of random matrices and many meta-parameters, and the feature vector of the NVAR was introduced equivalent to the readout of RC
For mathematical details, please refer to \cite{gauthier2021next}.
By applying NG-RC to three RC benchmark tasks including Lorenz attractor prediction, the model showed faster computational time while at the same time requiring only a small number of sample and few meta-parameters for training. 
A possible application would be using NG-RC to create a digital twin for dynamical systems.


\section{Recent applications of reservoir computing} \label{Recent Applications of Reservoir Computing}

As a special type of recurrent neural network, RC avoids the main problem of a difficult, unstable and resource-consuming training process. In the past decade, however, various deep learning algorithms, took advantage of the intricacies of gradient based RNN training with greater computational power and finally became main-stream. This led reservoir computing research into a niche for a few years. 
\textcolor{black}{As scientists in various research fields have found many new ways of RC implementations and applications (see Table \ref{tab:app}), RC have prompted renewed interest among researchers from disparate domains.
This section presents a comprehensive review of these recent trends, showcasing the widespread applicability of RC from the realms of engineering and computer science to the diverse fields of physical and social science.
}

\begin{table*}[]
\renewcommand{\arraystretch}{1.0}
\centering
\caption{Recent applications of RC in various research fields.}

\resizebox{0.85\textwidth}{!}{%
\begin{tabular}{lclll}
\hline \hline
 & \textbf{Research Fields}  &  & \multicolumn{1}{c}{\textbf{Applications}}                                                                                                                           &  \\ \cline{2-2} \cline{4-4}
 & Biomedical       &  & \begin{tabular}[c]{@{}l@{}}ECG, EMG and MCG signal processing \cite{hadaeghi2019reservoir, mastoi2019reservoir, n2021ecg, donati2018processing, sakib2021noise, mohsen2020ai}.\\ Medical images segmentation and classification \cite{hadaeghi2021spatio}.\end{tabular}         &  \\ \cline{2-2} \cline{4-4}
 &
  Machinery &
   &
  \begin{tabular}[c]{@{}l@{}}Active shape discrimination \cite{johnson2014active}.\\ Temporal information processing \cite{jiang2019physical}.\\ Robotic crawling \cite{bhovad2021physical}.\\ UAVs and sensors control \cite{chen2017caching, chen2019liquid, challita2019interference, bacciu2014experimental, palumbo2016human, sun2021sensor, konkoli2018developing}.\\ Fault diagnosis \cite{zheng2017brain, zheng2018fault, long2019evolving, zhang2019deep, zhang2021pre}.\end{tabular} &
   \\ \cline{2-2} \cline{4-4}
 & Data Science     &  & \begin{tabular}[c]{@{}l@{}}Series chunking and clustering \cite{atencia2019dynamic, atencia2020time, asabuki2018interactive}.\\ Similarity learning \cite{krishnagopal2018similarity}.\end{tabular}                                        &  \\ \cline{2-2} \cline{4-4}
 & Security         &  & \begin{tabular}[c]{@{}l@{}}Attack detection \cite{ozay2015machine, hamedani2017reservoir, hamedani2019detecting, chandrasekaran2021real}.\\ Specific emitter identification \cite{kokalj2020deep, kokalj2021reservoir}.\end{tabular}                                          &  \\ \cline{2-2} \cline{4-4}
 & Communications   &  & \begin{tabular}[c]{@{}l@{}}Optical communications \cite{wang2021signal}.\\ Network traffic flows analysis \cite{yamane2019application, ando2019road}.\\ Symbol detection in MIMO systems \cite{mosleh2018brain, zhou2020deep}.\end{tabular} &  \\ \cline{2-2} \cline{4-4}
 &
  Chemistry &
   &
  \begin{tabular}[c]{@{}l@{}}Remaining useful lifetime prediction of PEMFC \cite{morando2013fuel, morando2017anova, mezzi2018multi, li2020adaptive}.\\ Real-time gas concentration prediction- \\ using chemosensors \cite{fonollosa2015reservoir}.\\ DNA oscillators and Magnetic skyrmions \cite{goudarzi2013dna, yahiro2018reservoir, nguyen2020reservoir, liu2022reservoir}.\\ Chemical reaction networks \cite{yahiro2018reservoir, kan2021physical, nguyen2022sers}.\end{tabular} &
   \\ \cline{2-2} \cline{4-4}
 & Environmental    &  & \begin{tabular}[c]{@{}l@{}}Wind forecasting \cite{chitsazan2019wind, wang2020novel}.\\ Wind power generation \cite{hamedani2017reservoir}.\end{tabular}                                                    &  \\ \cline{2-2} \cline{4-4}
 & Audio and Speech &  & \begin{tabular}[c]{@{}l@{}}Speech recognition \cite{verstraeten2005isolated, ghani2010neuro, alalshekmubarak2014improving, zhang2015digital, abreu2020role}.\\ Music classification \cite{pons2019randomly}.\end{tabular}                                                   &  \\ \cline{2-2} \cline{4-4}
 & Finance          &  & \begin{tabular}[c]{@{}l@{}}Stock market prediction \cite{wang2021stock}.\\ Financial system modelling \cite{budhiraja2021reservoir}.\end{tabular}                                        &  \\ \hline \hline
\end{tabular}%
}
\label{tab:app}
\end{table*}

\subsection{Biomedical}

\begin{figure*}[ht]
    \centering
    \includegraphics[width=0.99\textwidth]{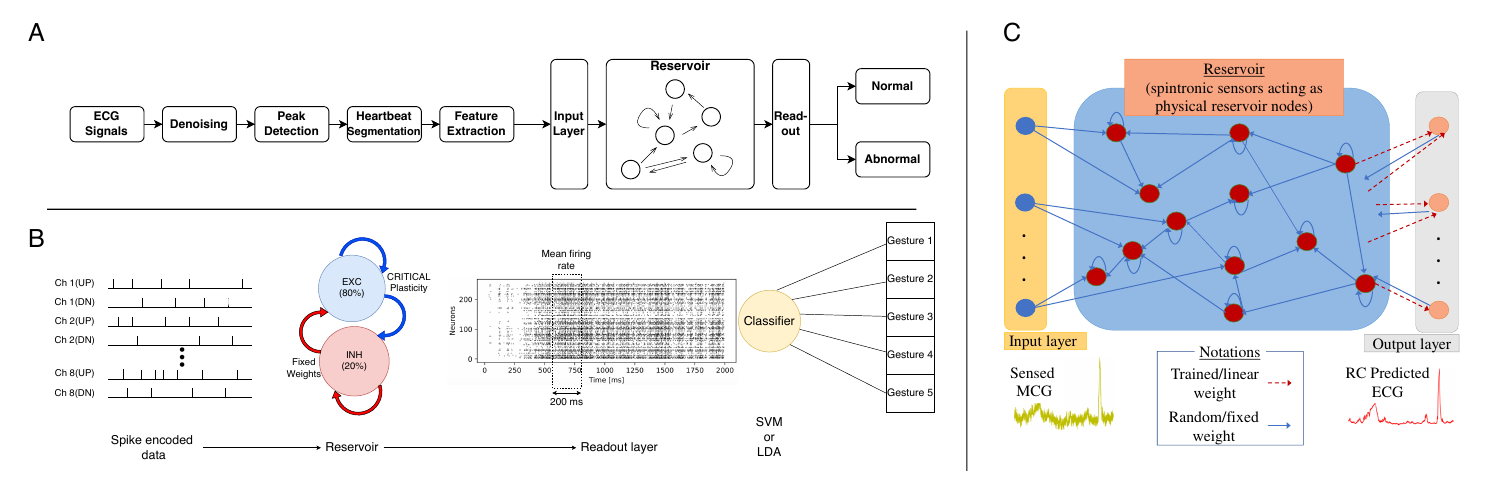}
    \caption{
    (A) Ventricular heartbeat classifier using RC \cite{mastoi2019reservoir}.
    (B) Gesture recognition using a plastic reservoir (EMG signal processing) by \cite{garg2021signals}.
    (C) MCG Monitoring using physical RC with spintronic sensors acting as physical reservoir nodes \cite{sakib2021noise}.
    }
    \label{fig:app_biomedical}
\end{figure*}

\textbf{Electrocardiogram (ECG).} 
A modified ESN is applied to \textit{cardiac monitoring} \cite{hadaeghi2019reservoir}. 
Specifically, the experimental data includes two classes of ECG signals from MIT-BIH databases with highly imbalanced number of instances. 
\textcolor{black}{The reservoirs are used as patient-adaptable classifiers. 
}
These classifiers can not only produce accurate results, but also show the potential to implement ECG classifiers by using neuromorphic hardware with spiking neural networks. 
Similarly, \cite{mastoi2019reservoir} used an ESN for a\textit{bnormal cardiac activity detection} (Fig. \ref{fig:app_biomedical}A). The main objective is to apply an ECG monitoring model in Medical Internet of Things (MIoT) devices with fast speed and low power consumption. The proposed RC model shows better performance and generalization in AHA and MIT-BIH-SVDM datasets than patient-adaptable methods, and also suggests that RC can be implemented in wearable wireless devices. 
Besides, in the problem of ECG signal \textit{denoising}, a single-node RC is applied to solve this problem by minimizing the EMG signal that impairs the ECG signal \cite{n2021ecg}.

In 2021, \cite{cucchi2021reservoir} reported a new type of reservoir for \textit{arrhythmic heartbeats} classification. Inspired by the pioneering work on the combination of organic electrochemical transistors (OECTs) and reservoir computing, they implemented dendritic networks using OECTs for real-time classification. In detail, the reservoir is created by using coupled dendritic fibers. 
Once these fibers are excited through the electrolyte, they create a strong enough non-linear dynamic of the input signals.
The proposed networks show the potential use for \textit{biofluid monitoring} and biosignal analysis with high accuracy, indicating that bio-compatible computational platforms can interact with body and biological analysis.

\textbf{Electromyography (EMG).}
Some studies on EMG also apply RC models.
\cite{kudithipudi2016design} proposed a scalable and reconfigurable neuromemristive reservoirs architecture for EEG and EMG signal analysis. 
A further EMG application using LSM model was proposed by \cite{donati2018processing}, where the EMG signal collected by the surface EMG (sEMG) sensors are classified by an LSM-based neuromorphic hardware. 
In 2021, another spiking RC model that applies CRITICAL plasticity rule \cite{brodeur2012regulation} for synaptic connection optimization was proposed for hand gesture recognition \cite{garg2021signals} (Fig. \ref{fig:app_biomedical}B). The author also proposed a novel approach to evaluate and convert the raw EMG signals to spikes encoding. 

\textbf{Magnetocardiography (MCG).} 
Checking the ECG is not possible for everyone. 
Alternatively, magnetocardiography (MCG) signals can be detected by measuring the magnetic field produced by the electrical currents in the heart and can be converted into ECG signals. 
\cite{sakib2021noise} built the first physical RC model for MCG monitoring. 
Specifically, the noisy sensed MCG signals take as input to the reservoir (i.e., the spintronic sensors, see Section \ref{Physical_RC}) and the output is the predicted ECG signals (Fig. \ref{fig:app_biomedical}C). 
This lightweight RC model is claimed to be much power-saving and lower memory-required while achieving comparable performance with a deep learning based filtering approach \cite{mohsen2020ai}.
A similar model was later proposed by \cite{shakya2021circuit} to extract ECG signals from MCG signals.

\textbf{Other biomedical applications of RC.}
In early research of molecular reservoir computing, the coupled deoxyribozyme oscillators is shown to be a type of reservoir \cite{goudarzi2013dna}. This refers to DNA reservoir computing (see Section \ref{Recent approaches in RC}).
In addition to DNA based RC, a medical image classification with distributed representations on cellular automata RC was reported by \cite{kleyko2017modality} (see cellular automata RC in Section \ref{Recent approaches in RC}).
Besides, a recent study on spatio-temporal feature learning used RC model for T‑cell consistent segmentation 
\cite{hadaeghi2021spatio}. Instead of only applying a single reservoir, the model used multiple reservoirs for image segmentation and classification, where each reservoir focuses on a specific area of the image to obtain local interactions.

\subsection{Machinery}
\begin{figure*}[ht]
    \centering
    \includegraphics[width=0.99\textwidth]{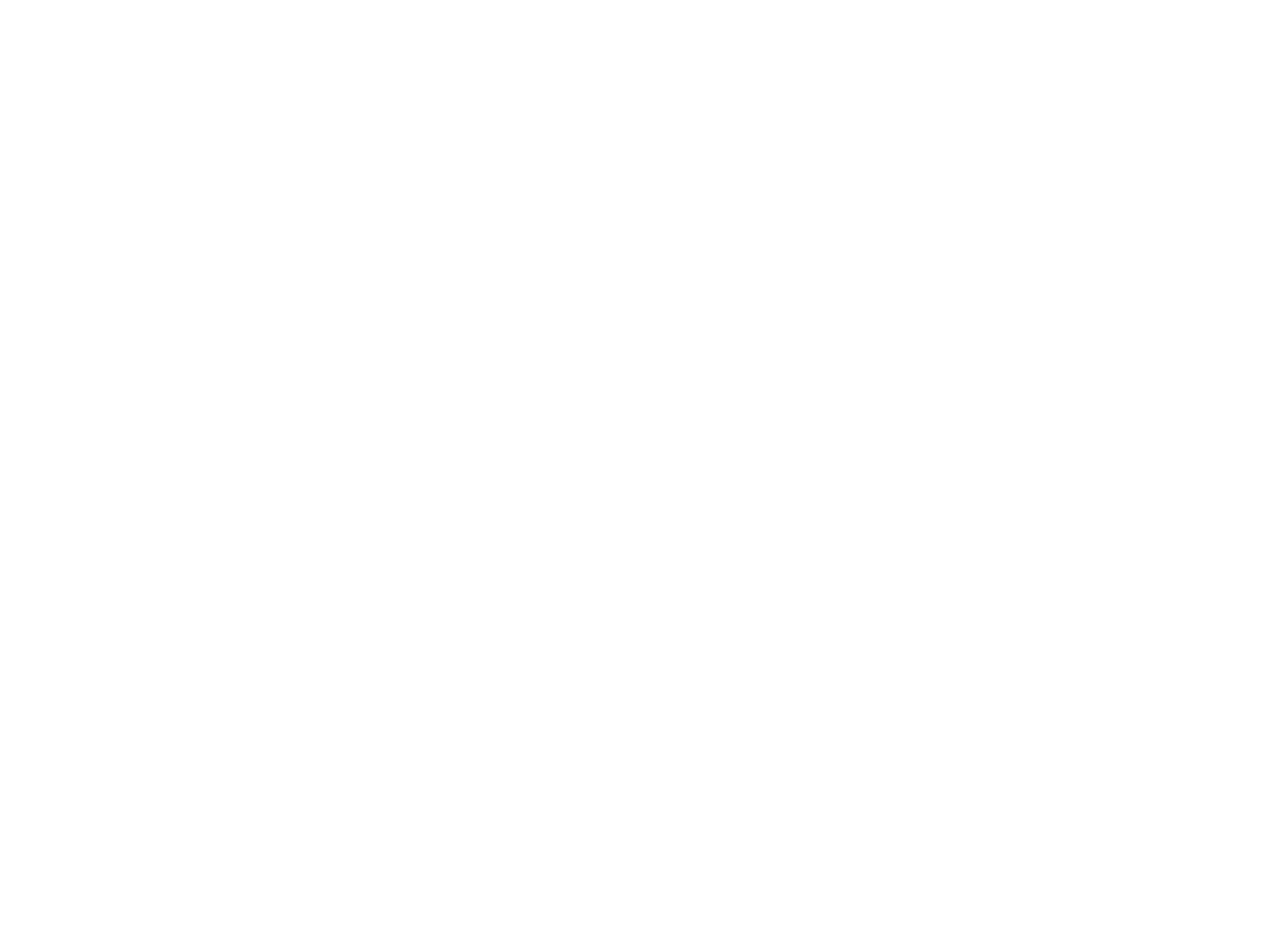}
    \caption{
    (A) Non-linear mass-spring-damper systems as a RC proposed by \cite{hauser2011towards} (middle panel). Pneumatically driven, modular robot arm as a RC proposed by \cite{eder2018morphological} 
    \textcolor{black}{(right panel)}.
    (B) A bio-inspired RC using octopus as soft robotic arm by \cite{nakajima2013soft}.
    (C) A foraging learning task \cite{yada2021physical}. A vehicle robot was placed in an environment with obstacles and was directed toward the goal. The system used FORCE learning to generate a coherent signal output from a living neuronal culture.
    }
    \label{fig:app_machinery}
\end{figure*} 

\textbf{Robotics.} 
In section \ref{Recent approaches in RC}, we gave a brief review of mechanical RC. 
The so-called mass-spring-damper systems were built as the first type of mechanical RC by \cite{hauser2011towards, hauser2012role} (see Fig. \ref{fig:app_machinery}A).
This new type of RC was then used for \textit{active shape discrimination}  \cite{johnson2014active}.
Meanwhile, \cite{nakajima2013soft} made a series of works on the octopus-inspired soft robotic RC (Fig. \ref{fig:app_machinery}B), showing that the implementations can learn to emulate timers, delays, and parity \cite{nakajima2013soft, nakajima2014exploiting, nakajima2018exploiting}.
Another application of robotic RC was reported to learn and reproduce various end point trajectories by using a new RC-based soft robot system with a highly complex \textit{pneumatically driven} robotic arm \cite{eder2018morphological}.
\textcolor{black}{
RC has also been applied in robotic crawling by using the \textit{origami} -- a traditional play of folding paper into sophisticated and 3D shapes. \cite{bhovad2021physical} shows that an origami structure based PRC can be designed to build a soft robotic controller for earthworm-like peristaltic crawling. 
}
\cite{yada2021physical} embedded FORCE learning into a robot to learn in a living neuronal culture (i.e., foraging learning task), in which the robot was placed on square fields with various obstacles and was directed toward the target objects (see Fig. \ref{fig:app_machinery}C).

\textbf{Unmanned Aerial Vehicles (UAVs).} 
RC can be applied to UAV systems. 
Particularly in telecommunications, RC models were used in the so-called cache-enabled UAVs for optimizing \textit{resource allocation} over the LTE licensed and unlicensed bands.
The first attempt was to use ESN in such a system to predict each user’s content request distribution and its mobility pattern when limited information on the states of users and the network is available \cite{chen2017caching}.
An LSM-based model was further proposed, which can predict more context information of the users and thus improves the prediction accuracy \cite{chen2019liquid}.

\begin{figure*}[ht]
    \centering
    \includegraphics[width=0.99\textwidth]{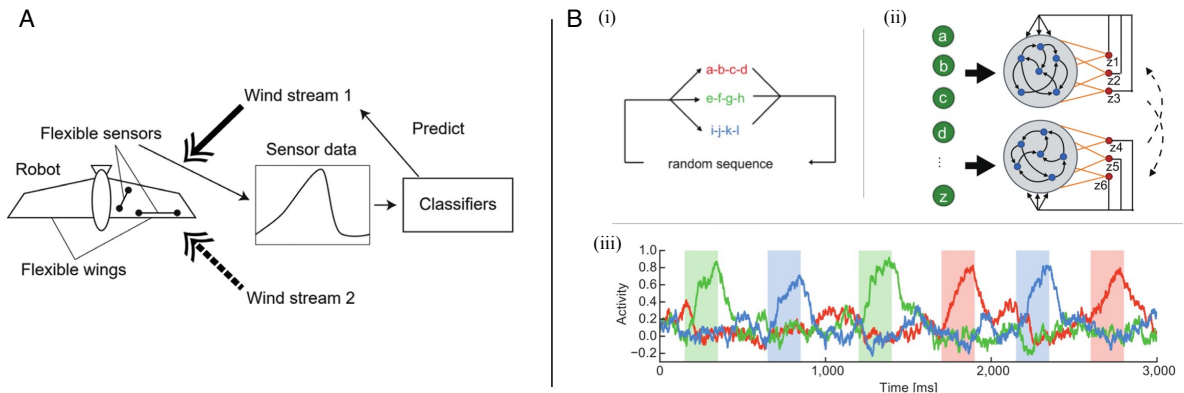}
    \caption{
    (A) A flapping-wing UAV for recognition of the wind direction in which PRC approach was used to classify the wind stream from sensors \cite{tanaka2021flapping}.
    (B) A dual-reservoir structure for chunking temporal information streams \cite{asabuki2018interactive}.
    (i) Chunking problems example. Three fixed chunks and noise (random sequence between chunks) are repeated in the input sequence with equal probabilities. (ii) The dual-reservoir structure. (iii) Selective readout responses to the individual chunks, which are colored according to their selectivity to the chunks.
    }
    \label{fig:app_datasci}
\end{figure*}

Another branch of UAV applications includes one that uses a deep ESN based reinforcement learning algorithm for UAV path planning by \cite{challita2019interference}. In this system, each UAV uses an ESN to optimize paths and learns transmission power at different locations.
Besides, ESN can also be used to control rotorcraft UAVs, which outperforms linear models in robustness to disturbances \cite{vargas2015swing}.
In 2021, \cite{tanaka2021flapping} proposed a method for controlling flapping-wing UAVs in different wind directions, where strain sensors are applied to measure the wind movements, and a physical RC is used as a classifier to recognize the wind stream from the sensor data (see Fig. \ref{fig:app_datasci}A).

\textbf{Sensors.}
RC models can also be integrated into wireless sensor networks (WSNs). 
Generally, the sensor devices in WSNs are distributed and computationally constrained, and the collected data usually consist of temporal information, which makes RC inherently suitable for embedding on the WSN devices \cite{bacciu2014experimental}.
One of the real-world WSN applications using RC is activity recognition in Ambient Assisted Living (AAL) tasks \cite{palumbo2016human}.
Specifically, RC-based multi-sensors were used for feature collection and extraction. 
The sensed data were then further processed by an ESN which provides a good activity recognition accuracy with low computational costs. 
In 2021, a bio-inspired in-sensor RC was demonstrated to be effective for classifying short sentences of language \cite{sun2021sensor}. 

It is worth noting that although RC had been shown potential for processing sensor data, some researchers like \cite{konkoli2018developing} argued that
those reservoirs focusing on sensing are often exploited in a somewhat passive manner, being a separated post-processing component that receives data from sensors. 
Therefore, \cite{konkoli2018developing} further proposed the State Weaving Environment Echo Tracker (SWEET) sensing approach. Here, RC was considered as the sensing element itself for novel sensing applications such as \textit{ion} concentration analysis. 

\textbf{Fault Diagnosis.} 
Fault diagnosis generally refers to the process of detecting errors in physical systems while attempting to identify the source of the problems.
Built on the deep ESN architecture suggested by \cite{gallicchio2017deep}, \cite{long2019evolving} proposed evolving deep ESN models for 3-D printer fault diagnosis, with a developed version of particle swarm optimization (i.e., competitive swarm optimizer, CSO). This RC model uses evolutionary optimization and is shown to be state-of-the-art and computationally economic, which is a complement to deep learning algorithms, rather than a competitor. 
Meanwhile, the same research group proposed another solution for 3-D printer fault diagnosis \cite{zhang2019deep, zhang2021pre}. Specifically, deep ESNs were used to improve feature extraction performance, where the features were reinforced throughout the hidden layers by using fuzzy clustering as a tuning step.
This low computational costing model also provides the optimal solution in all experiments, with a total of 26 different condition patterns in fault diagnosis data.

In addition, RC can also be applied to chemical fault diagnosis in 
the \textit{proton exchange membrane fuel cell} (PENFC) system. 
The first RC application in PEMFC system diagnosis was made by \cite{zheng2017brain}, where a delayed feedback RC was used to detect four fault types yet in the static operating conditions only. 
Instead of processing voltage signal in the original data space, a newer variant based on previous models was proposed, which performs abnormal detection in the reservoir computing based model space (current-voltage model) without requiring additional feature extraction \cite{zheng2018fault}.

\subsection{Data Science}

\textbf{Image recognition.}
Image recognition stands as a prominent field within computer vision and machine learning.
While RC shows impressive performance in finding and generating temporal features, it has also been adapted for image recognition tasks, either alone or integrated with other techniques such as deep networks with convolutional layers.
Table \ref{tab:performance} shows the performance comparison between different RC models.
To illustrate, when using RC alone to deal with 2D or 3D inputs (i.e., images or videos), data are usually flattened into 1D signals prior to feeding them into the reservoir, as discussed in \cite{lukovsevivcius2012practical}. 
When combining with convolutional layers, these layers preprocess images and videos, transforming them into intermediate representations which the reservoir can process temporally
\cite{jalalvand2018application, tong2018reservoir, tanaka2022reservoir, chang2020reinforcement, chang2019convolutional, yoshihiro2020}.
Although RC models demonstrate competitive performances with other machine learning methods like CNNs on simpler datasets such as MNIST, their performance significantly declines when faced with datasets exhibiting higher spatial complexity, such as CIFAR-10. 
This performance gap highlights the inherent limitations of RC in dealing with complex spatial correlations, and underscores the need for further exploration and investigations in this field.
On the other hand, studies have proved that convolutional neural networks are likely to misclassify even if small perturbations are added to original samples \cite{goodfellow2014explaining, moosavi2017universal, tsipras2018robustness, su2019one, kotyan2020evolving, kotyan2022adversarial}. 
Thus, this also highlights a strong incentive for RC-based methods to tackle high-dimensional inputs with strong 2D/3D correlations, as it was shown that higher degrees of nonlinearity in the model are related to more robust neural networks, and nonlinearity is where the RC really shines.

\begin{table}[]
\renewcommand{\arraystretch}{1.0}
\centering
\caption{
\textcolor{black}{Comparison of the \textit{Accuracy} (\%) of recent RC models in image recognition benchmarks.}
}
\resizebox{0.6\linewidth}{!}{%
\begin{tabular}{lclrrr}
\hline \hline
\multicolumn{1}{c}{\textbf{Model Type}} &
  \textbf{Model} &
   &
  \multicolumn{1}{c}{\textbf{MNIST}} &
  \multicolumn{1}{c}{\textbf{Fashion MNIST}} &
  \multicolumn{1}{c}{\textbf{CIFAR-10}} \\ \hline \hline
\textbf{Standard RC}                       & \cite{schaetti2016echo} &  & 99.07\% & -       & -       \\ \cline{1-2} \cline{4-6} 
\multirow{4}{*}{\textbf{Hybrid (CNN+RC)}}  & \cite{jalalvand2018application} &  & 99.19\% & -       & -       \\
                                           & \cite{tong2018reservoir} &  & 99.25\% & -       & -       \\
                                           & \cite{yoshihiro2020} &  & 98.71\% & 86.27\% & -       \\
                                           & \cite{tanaka2022reservoir} &  & 98.38\% & 91.04\% & 64.49\% \\ \cline{1-2} \cline{4-6} 
\multirow{4}{*}{\textbf{Physical RC}}              & \cite{jacobson2021image} &  & 97.70\% & -       & -       \\
                                           & \cite{tran2019hierarchical} &  & 73.86\% & -       & 12.96\% \\
                                           & \cite{yang2022optical} &  & 98.20\% & 89.90\% & -       \\
                                           & \cite{moran2018reservoir} &  & 98.08\% & -       & -       \\ \cline{1-2} \cline{4-6} 
\multirow{2}{*}{\textbf{Hybrid (CNN+PRC)}} & \cite{jacobson2021hybrid} &  & 98.90\% & -       & -       \\
                                           & \cite{an2020unified} &  & 99.03\% & -       & 60.57\% \\ \hline \hline
\end{tabular}
}
\label{tab:performance}
\end{table}

\textbf{Clustering.} 
One of the applications of RC in data science is \textit{clustering}. As a special case of clustering, \textit{time-series clustering} introduces several additional issues when compared with static data clustering. 
For example, the lengths of time-series usually vary, and some of them may be infinite (e.g., video and audio sequences collected from CCTV cameras). 
Moreover, temporal dependencies in different parts 
of a particular time-series cannot be captured by making a fixed detecting window (i.e., the dynamical behaviors in sequences contain to both short- and long-term correlation).
As a result, similarity measurement techniques, such as calculating Euclidean distance among temporal data, are not inherently suitable for time-series clustering. 

\cite{atencia2019dynamic} proposed the first dynamic clustering algorithm using conventional ESNs. The idea is to apply a clustering method inside every step of the reservoir's state update, where the author claimed that any unsupervised clustering methods can be used in principle (e.g., k-means or any other iterative, partitioning clustering methods). 
The proposed method overcomes the above-mentioned issues and produces more compact clusters when applied to a hard classification problem of detecting patients with eye disease in eye movements datasets (saccades). 
In 2020, deep reservoir structure was introduced in time-series clustering \cite{atencia2020time}. 
The proposed algorithm was applied to more common benchmark datasets and showed better clustering quality than the previous algorithm and static clustering methods.

\textbf{Chunking.} 
Related to clustering problems, some studies have been proposed for sequence \textit{chunking} \cite{vargas2021continual, asabuki2020somatodendritic}. Here, the main difference between time-series clustering and sequence chunking is that chunking finds the temporal correlation between state variables, instead of clustering homogenous time-series together based on a certain similarity measure (see Fig. \ref{fig:app_datasci}B in the upper-left panel).
\cite{asabuki2018interactive} used dual-reservoir networks that supervise each other to mimic the partner's responses to the given input. Here, a challenge of chunking sequences with uniform transition probabilities, which can be easily processed by humans in basal ganglia, was successfully solved by the proposed model while conventional statistical approaches fail to chunk (see Fig. \ref{fig:app_datasci}B). This suggests that reservoirs can predict dynamical response patterns to sequence input other than to directly learn transition patterns. 

\textbf{Similarity learning.} 
\cite{krishnagopal2018similarity} applied RC to learn the similarity between image pairs with limited data.
The reservoir here acts as a non-linear filter that projects the images into a high-dimensional state space, in which the state trajectories represent different dynamical patterns that reflect the corresponding relationship of given image pairs. 
The proposed model was tested on MNIST dataset and images taken from a moving camera. Compared to deep Siamese Neural Networks, this RC model showed significantly better performance in generalization tasks. 
The generalized combinations of relationships provide robust and effective image pair classification.


\subsection{Security}
\begin{figure*}[ht]
    \centering
    \includegraphics[width=0.99\textwidth]{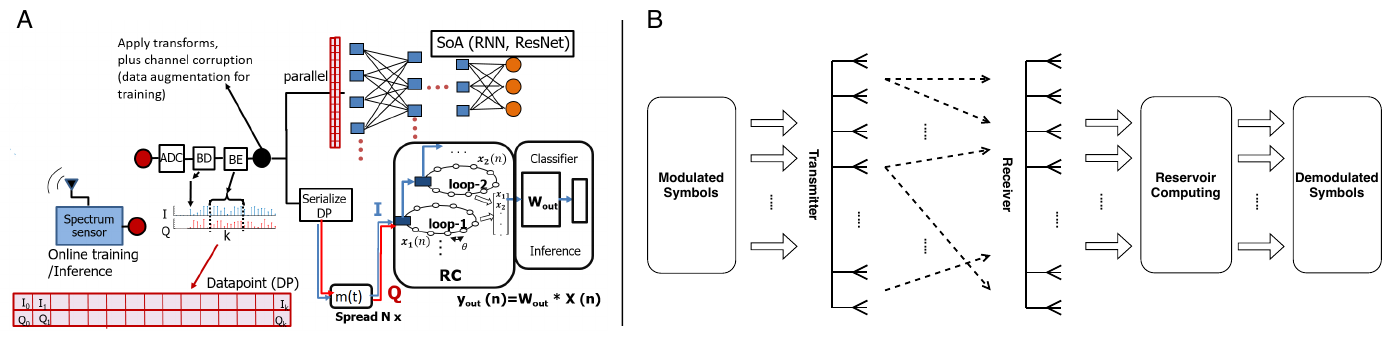}
    \caption{
    (A) Diagram of Specific Emitter Identification (SEI) using RC \cite{kokalj2021reservoir}. A multi-layer time-delayed feedback reservoir structure was introduced to enable the linear classifier for emitter identification.
    (B) RC used for symbol detection (modified from \cite{hamedani2019novel}). 
    }
    \label{fig:app_security}
\end{figure*}

\textbf{Attack detection in Smart Grid.} 
RC has been proven to efficiently solve \textit{false data injection} (FDI) and to improve the reliability in smart grid systems \cite{ozay2015machine}. 
The first attempt was to use a modified delayed feedback network (i.e., single-node time-delayed RC) as a reservoir combined with a multilayer perceptron (MLP) as a readout for \textit{single-period attack detection} \cite{hamedani2017reservoir}. 
This RC model with MLP architecture produces a high attack detection rate (99\%) and shows strong robustness in various attack types.
Later, the author extended the pioneering work to the more challenging dynamic attack detection in smart grid \cite{hamedani2019detecting}, where a bio-inspired learning rule called \textit{precise-spike-detection} (PSD) \cite{yu2013precise} is used for spiking reservoir training.

Regarding the attack detection, recent applications of RC include detecting malware and micro-architectural attack, which is reported in \cite{chandrasekaran2021real} using a CMOS-based RC neural network embedded in a 65nm CMOS chip.

\textbf{Specific Emitter Identification (SEI).}
SEI is capable of extracting rich non-linear characteristics of internal components within a transmitter to distinguish one transmitter from another. 
Since the fingerprint of SEI cannot be emulated, it is widely used in IoT devices to prevent MAC address spoofing attacks.
A reservoir with delay loops for SEI was first proposed by \cite{kokalj2020deep} and further adapted 
to edge computing \cite{kokalj2021reservoir} where the RC architectures include a digital loop (FPGA) and a photonic one (Fig. \ref{fig:app_security}A).

\subsection{Communications}
\textbf{Optical communications.} 
In the high speed optical fiber communication systems, RC was applied for \textit{digital equalization}. \cite{wang2021signal} quantified the equalization performance of the optoelectronics RC. Experiment results show that the optoelectronics RC outperforms traditional equalizers under the same transmission conditions, taking the advantage of its ring topology for better correlation between adjacent data as well as its lower complexity and computational cost.  

\textbf{Network traffic.}
\cite{yamane2019application} proposed a method for application identification for network traffic by physical RC, which processes traffic flows as dynamical time series data and enables fast and real-time identification.
Another RC application is reported by \cite{ando2019road} for road traffic analysis.

\textbf{Symbol detection in MIMO-OFDM systems.} 
In wireless communication domains, multiple-input multiple-output with orthogonal frequency division multiplexing (MIMO-OFDM) is a key enabling technology in the 5G cellular network. 
Symbol detection is an important technique due to the severe non-linear distortion during transmission (Fig. \ref{fig:app_security}B). 
Thus, an accurate estimation of MIMO-OFDM channel is usually required.
The first integration of RC and MIMO-OFDM systems was proposed by \cite{mosleh2018brain}. Specifically, an ESN was used for system modelling and predicting non-linear dynamics, where the MIMO-OFDM channel estimation is no longer necessary. 
Further, inspired by deep RC architectures \cite{gallicchio2017deep}, \cite{zhou2020deep} extended the existing shallow RC to form a deep neural network \cite{zhou2020deep, hamedani2019novel}, which significantly mitigates the frequency distortion.

\subsection{Chemistry}

\begin{figure*}[ht]
    \centering
    \includegraphics[width=0.99\textwidth]{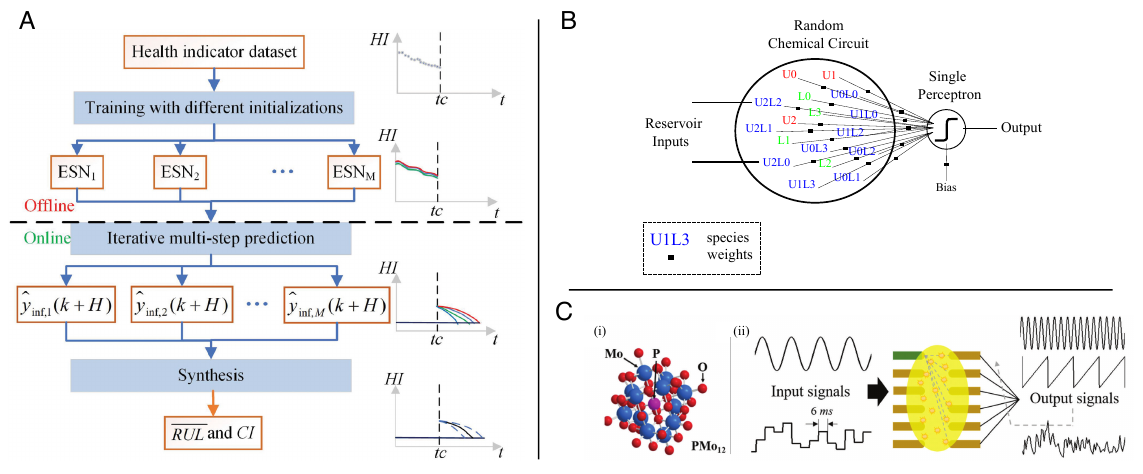}
    \caption{
    (A) An ESN was used to predict the remaining useful lifetime (RUL) of Fuel Cells \cite{li2020adaptive}.
    (B) A chemical RC implemented by a random chemical circuit with different DNA species 
    \cite{nguyen2020reservoir}.
    (C) An electrochemical-reaction-based reservoir proposed by \cite{kan2021physical}.
    (i) Structure of the Polyoxometalate (POM) molecule. 
    (ii) Testing procedure.
    }
    \label{fig:app_chem}
\end{figure*} 

\textbf{Fuel Cells (FC).} 
ESNs have shown their effectiveness on \textit{remaining useful lifetime} (RUL) prediction for \textit{proton exchange membrane fuel cell} (PEMFC). \cite{morando2013fuel} developed the first RC model for FC prognostics using conventional ESN. 
Later, several variants of RC model have been proposed for improving the prediction performance. 
These include an ESN combined with ANOVA method \cite{morando2017anova}, as well as a multi-reservoir ESN \cite{mezzi2018multi}. 
However, recent research states that the above proposals assume that FCs are operated in constant nominal operating conditions \cite{li2020adaptive}; that is, only the degradation is considered the factor of the deviation of stack voltage. 
Another open problem is that the prognostic results in long-term experiments show that prediction will become inaccurate when disturbances occur.

\textbf{Chemosensor.} 
Metal oxide (MOX) based sensors are a common choice for tasks of chemical detection, yet the time response of these chemical sensors is usually excessively slow. 
It is stated that algorithms based on batch or sequential measurements are not suitable for continuous sensing scenarios. 
\cite{fonollosa2015reservoir} used RC algorithms to overcome the slow temporal dynamics of the chemosensors and applied RC for real-time gas concentration prediction by observing the sensors’ time series in response to the changes in the composition of a gas sample. 
Still, problems were reported, such as the drift of sensor response over time.

\textbf{Magnetic skyrmions.} 
RC using magnetic \textit{skyrmions} are reviewed previously in Section \ref{Recent approaches in RC}, where the random phase structures of the skyrmion fabrics are suitable for RC implementation \cite{prychynenko2018magnetic}. 
Recent applications in this field is to implement RC based on a single magnetic skyrmion memristor (MSM) for image classification task (i.e., handwritten digit recognition) \cite{jiang2019physical}.

\textbf{Coupled deoxyribozyme oscillators and DNA oscillators.}
Taking the inspiration of DNA reservoir computing approach \cite{goudarzi2013dna}, \cite{nguyen2020reservoir} proposed a random chemical RC model, where the random chemical circuits (i.e., DNA strand displacement) provide complex non-linear dynamics, making them suitable for RC implementation (Fig. \ref{fig:app_chem}B). 
The proposed model outperforms the previous deoxyribozyme oscillator RC \cite{yahiro2018reservoir} in short and long-term memory tasks. 
Another recent RC using DNA oscillators was reported in \cite{liu2022reservoir} which solves the problem of the lack of readout layer \cite{goudarzi2013dna}, 
and then applied to a handwritten digit recognition and a second-order non-linear prediction task.

\textbf{Chemical Reaction Networks (CRNs).}
As mentioned earlier, reservoir's dynamic can be generated by a set of ordinary differential equations (ODEs).
An extension was proposed for single stranded DNA (ssDNA) analysis \cite{nguyen2022sers}.
Besides, \cite{yahiro2018reservoir} used a modular framework for molecular computing to implement a RC model. 
The main advantage of this work, compared with previous DNA oscillators \cite{goudarzi2013dna}, is that molecular computing allows tuning the size of CRNs.
Another new chemical RC architecture was proposed by \cite{kan2021physical}, where the reservoir is implemented through electrochemical reactions since the chemical dynamic is shown to be computing resources (Fig. \ref{fig:app_chem}C). 
As claimed by the author, the Polyoxometalate molecule in the solution ``increases the diversity of the response current and thus improves their abilities to predict periodic signals''.

\subsection{Environmental}
\textbf{Wind forecasting and wind power generation.}
\cite{chitsazan2019wind} proposed a RC based wind speed and wind direction forecasting model. In fact, the proposed model is a new type of non-linear echo state network, which is discussed in Section \ref{Recent approaches in RC}. 
Instead of deterministic forecasting, a recent study 
focuses on probabilistic wind power forecasting \cite{wang2020novel} by using a time warping invariant echo state network \cite{lukovsevicius2006time}.
In addition, the wind turbines were used as a major source of power generation for smart grids, where the delayed feedback RC was applied for attack detection \cite{hamedani2017reservoir}.


\subsection{Audio and Speech}

\textbf{Audio processing.} 
Audio signals cover a wide range of temporal sequences such as speech, sounds and music. 
In 2009, RC was first applied as a general framework for non-linear audio processing by \cite{holzmann2009reservoir}. 
Three main potential applications were proposed with simulations, including tube amplifier plugin identification, non-linear audio prediction and music information retrieval (MIR). 
RC was claimed suitable for non-linear audio processing because of the inherently temporal processing capability. 
In terms of real-time audio processing, a cascaded discrete-time RC was proposed for black-box system identification \cite{keuninckx2017real}. Albeit much effort was made to reduce computation consumption, the cascaded structure is stated considerably more complex to tune than the conventional RC.
In 2019, \cite{pons2019randomly} proposed randomly weighted CNNs for music classification. 
The proposed model shares similarity to RC, where weight connections remain untrained during training.


\textbf{Speech recognition.} 
The earliest RC application of speech recognition was presented in \cite{verstraeten2005isolated}. Here, an LSM-based RC with spiking integrate-and-fire neurons was implemented recognizing isolated digits, where the readout is trained using ridge regression. 
The problem of the model is that it requires intermediate data storage for offline learning.
Similarly to the work above, \cite{ghani2010neuro} trained output neurons using back-propagation based MLPs.
Inspired by Hebbian learning, \cite{zhang2015digital} further proposed a variant of Hebbian online learning rule to train an LSM without requiring data storage for speech recognition. 
In detail, the analog input speech signal is pre-processed by the \textit{Lyon passive ear} model and further converted into spikes by BSA algorithm \cite{schrauwen2003bsa} before feeding into LSM. 
Other types of RC can also be applied to speech recognition. 
For example, an ESN combined with extreme kernel machines was used for Arabic speech recognition \cite{alalshekmubarak2014improving}. 
In 2020, a RC based on nano-oscillators was also applied to TI-46 database \cite{abreu2020role}.

\subsection{Finance}
\textbf{Stock market prediction.}
A successful prediction of a stock's future price could yield significant profit.
An early attempt of short-term stock price prediction was reported in \cite{lin2009short}, who used an ESN as a basic network with the \textit{Hurst exponent} to select a persistent subseries with the greatest predictability for training from the original training set.
Instead of using basic ESN, three RC network structures were investigated in stock price prediction \cite{wang2021stock}, including the small-world topology discussed in an earlier section.

\textbf{Financial System modelling.} 
\cite{budhiraja2021reservoir} used RC for financial system modelling. In this study, an ESN was first applied to predict a pre-defined financial system behavior. The model was further proved to effectively re-generate only the required data based on limited known information.



\section{RC with Brain Mechanisms and Cognitive Science}
\label{RC with brain mechanisms and cognitive science}

\subsection{Reservoir in the Cerebral Cortex}
RNNs have been shown to have rich, complex, non-linear and high-dimensional dynamics. 
In the cerebral cortex, especially the prefrontal cortex (PFC), massive recurrent connections of neurons were found, 
and it is progressively recognized that some parts of the brain operate as reservoirs \cite{mante2013context}. 
Moreover, the cortex is shown able to extract the desired outputs (readout) from the high-dimensional neural representations (reservoir). 
In this section, we review studies on using RC to model the cerebral cortex.

\subsubsection{Dominey’s Decade-long Research: the Birth of RC} 
Dominey et al. developed the first RC prototype in a series of neurocognitive studies on corticostriatal systems \cite{dominey1995complex, dominey1995model}. 
During the period of 80s-90s, many researchers were focusing on the characterization of the fast eye movements (i.e., the oculomotor saccade) in the corticostriatal system, which refers to the interactions between cortex and basal ganglia \cite{bruce1985primate}. 
Particularly, \cite{barone1989prefrontal} examined the function of the corticostriatal system by carrying saccade experiments on macaque monkeys.
The experiments showed that some neurons 
(1) have a preferred spatial saccade amplitude and direction; 
(2) selective to response to a particular sequential order. 
In 2013, \cite{rigotti2013importance} characterized this finding as \textit{mixed selectivity}, which became one of the important principles in RC and cognitive science.

\begin{figure*}[ht]
    \centering
    \includegraphics[width=0.99\textwidth]{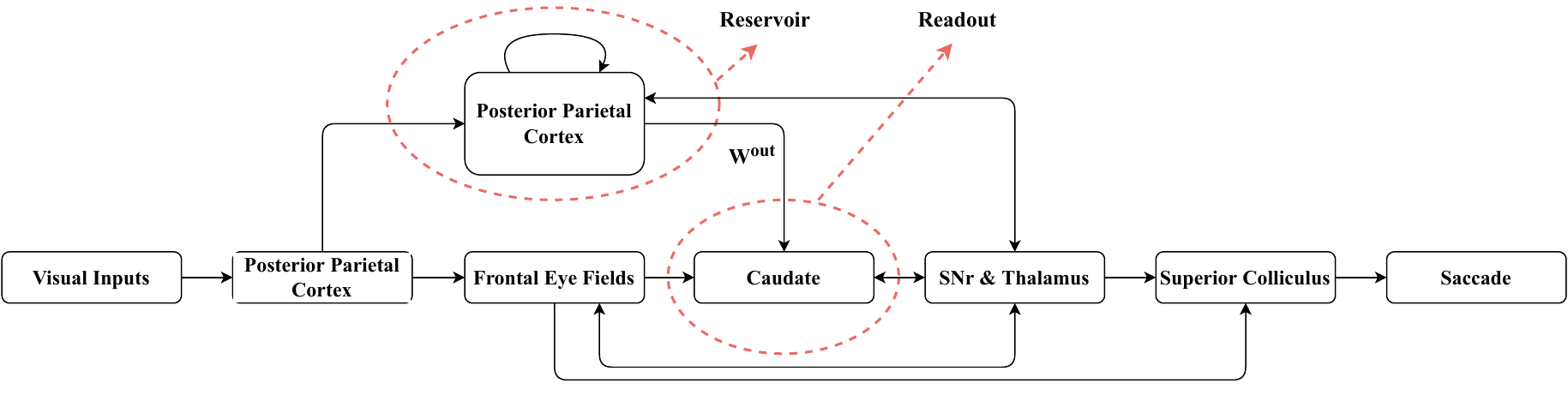}
    \caption{
    The first RC model built based on a recurrent PFC system as reservoir
    (modified from \cite{dominey1995model}).
    }
    \label{fig:cog_PFC}
\end{figure*}

Suggested by the two experiments of the \textit{corticostriatal} saccade system \cite{barone1989prefrontal, dominey1992cortico},
the first corticostriatal RC model was built based on (1) a recurrent prefrontal cortex (PFC) system (i.e., the reservoir), and (2) the reward-related learning in PFC-to-caudate connections (i.e., the readout).
Fig. \ref{fig:cog_PFC} shows the architecture of the model.
Since they found that the modification of the recurrent connections are considerably computational costing, they decided to initialize PFC layer (the reservoir) with a mixture of fixed inhibitory and excitatory recurrent connections. 
The reservoir layer was then connected to the \textit{caudate} or \textit{striatum} to obtain the readout. This pioneering RC model, as the author claimed, can be seen as a dedicated temporal recurrent network (TRN), which shows the inherent capabilities and sensitivity to temporal and sequential structure by providing a rich spatio-temporal dynamic \cite{dominey2000neural}. 

\textbf{Recent extension of TRN.} Dominey et al. further proposed a series of works on the previous corticostriatal RC model. These include a combination of RC and neuro-physiological models of language processing \cite{dominey2009cortico, dominey2009neural}, as well as a performance improvement of the RC learning algorithm \cite{hinaut2013real}. Reader may refer to a more detailed review of the corticostriatal RC model and its historical developments in a review paper by \cite{dominey2013recurrent}.
In 2013, one of the important cortical activities was obtained in randomly connected recurrent networks (e.g., reservoir) and was then characterized as \textit{mixed selectivity} \cite{rigotti2013importance} (see the following section). 
In 2016, the representational power and dynamical properties of mixed selectivity were investigated by training a RC model to perform a complex cognitive explore-exploit task initially developed for monkeys \cite{enel2016reservoir}. By comparing the neural activity of the reservoir and the primate dACC neurons, it is found that not only mixed selectivity was observed in the two types of neurons, but more strikingly, the distributions of neurons were quite similar in terms of the epoch (explore/exploit), the task phase, and the target choice, which strongly supports the argument that the cortex behaves computationally as a reservoir (see Figures 3 and 4 in \cite{enel2016reservoir}).

\subsubsection{Recent Studies of the Cortex}
\textcolor{black}{\textbf{Dimensionality implies selectivity.}} 
One of the complex neural activity phenomenons in PFC and in the model by \cite{dominey1995complex} is that the firing rates of some neuron populations were modulated by the combinations of conditions such as spatial location and sequential order \cite{dominey2021cortico}. 
This cortical activity was then characterized as mixed selectivity by \cite{rigotti2013importance} in an object sequence memory task.
Specifically, pure selectivity refers to neurons whose responses are selective only to an individual task-relevant aspect, whereas mixed selectivity refers to neurons whose responses are explained by a non-linear superposition of responses to the individual task-relevant aspects. 
In the object sequence memory task, monkeys were required to watch a sequence of two subsequently displayed images. After that, they had to 
(1) recognize the two images under distraction (recognition task), or
(2) recall the order of the two images (recall task). 
The \textit{dimensionality} of the neural spaces was then estimated (i.e., the minimal number of coordinate axes needed to specify the position of all points in neurons' firing rate space).
It is observed that the dimensionality was higher if neurons having mixed selectivity were included. 
More importantly, the neural population was estimated to have a higher dimensionality when the monkeys performed correctly on a trial. That is, in the error trials, a collapse in dimensionality was observed, which impairs the ability of downstream readout neurons to produce the correct response. 
Moreover, \cite{rigotti2013importance} and \cite{fusi2016neurons} showed that RC models can be compared with a randomly connected recurrent structure in the monkey prefrontal cortex, which can generate high-dimensional mixed selective dynamics to assure the separability in the downstream readout units. 
The higher the dimensionality of the population coding, the better the performance on the task \cite{dominey2021cortico}. 
Regarding the readout, it was reported that the brain implements mixed selectivity even when it does not enable behaviorally useful linear decoding (i.e., simple linear readout), suggesting that mixed selectivity may be the key of population encoding for reliable and efficient neural representations \cite{johnston2020nonlinear}.

Recent studies have shown that mixed selectivity not only plays an important role in PFC, but in other parts of the brain. 
\cite{ledergerber2021task} found strong mixed selectivity in the subiculum (i.e., the area between the entorhinal cortex and the CA1 subfield), where individual neurons respond conjunctively to task-related aspects including position, head direction, and speed.
In 2022, mixed selectivity was observed in the thalamus of a weakly electric fish \cite{wallach2022mixed}. Here, the mixed selectivity strategy was implemented to encode interactions in the recurrent networks in pallium, which is related to courtship and rivalry in terms of dominance in male-male competition and female-mate selection.




\subsection{Neuronal Oscillations}
Neuronal oscillations refer to the temporally structured activity generated in mammalian brains, where neurons undergo periodic changes in excitability. These oscillations had been found in neuron assemblies, a concept to describe the behaviors by a population of neurons. 
In this section, we first present the cognitive science research on neuronal oscillations, and then we discuss the relationship between oscillations and RC models, followed by several examples of the existing research. Contents are partially from \cite{singer2021cerebral}.

\subsubsection{Neuron Assemblies}
\textbf{Feed-forward circuits.}
It is widely believed that there are two frameworks of processing in natural systems in the cortex \cite{singer2021cerebral}: 
(1) convergent feed-forward circuits and 
(2) neuronal assemblies. 
In the framework of feed-forward circuits, specific neurons fire to particular features, and the information is propagated from the former layer to the next higher layer. 
In this way, higher-level features (e.g., cognitive objects) are extracted through a multi-layered structure. 
The encoding scheme here refers to spatial encoding, which is well-suited for simultaneously presenting features such as images. 
However, due to the lack of short-term memory functions, feed-forward circuits are less apt to tackle the relations among temporally segregated events.

\textbf{Neuron assemblies.}
On the other hand, a complementary framework in cognitive brains is the neuronal assemblies. 
Unlike the feed-forward networks which include explicit layered structures, the assemblies of neurons usually form coupled recurrent networks with non-linear, high-dimensional and self-organizing dynamic
\cite{singer2021cerebral}.
Relations among cognitive objects are translated into the weighted connections between neurons; in other words, high-level features are represented by the amplified reverberations (echoing) of neuronal assemblies.
With the reverberating responses, the rich dynamics provided by the coupled recurrent connections have short-term memory (fading memory), and become efficient to handle temporally related sequential events.

\subsubsection{Blinding Problem}
\textbf{Problem of neuronal assemblies.}
One of the challenging problems of the neuron assemblies is the blinding problem, which refers to the segregation of simultaneously active assemblies.
According to \cite{hebb2005organization}, if assemblies were
solely distinguished by enhanced activity (i.e., discharge rate) of the constituting neurons,
it becomes difficult to distinguish which of the more active neurons actually belong to which assembly. Moreover, if the given objects share some common features and overlap in space (e.g., blind source separation and cocktail party problem), the corresponding feature-selective nodes would have to be shared by several assemblies \cite{singer2021cerebral}.

\textbf{Solutions to blinding problem.}
A possible solution is multiplexing, in which various active assemblies are segregated in time. Because of the discharge rate of cortical neurons is relatively low (i.e., the integration needs time), multiplexing becomes problematic if only discharge rate is considered for distinguishing assemblies \cite{tovee1992functional}.
Therefore, it is only capable in a slow timescale \cite{vanrullen2005spike}.

In the 1990s, Gray and Singer proposed that ``neurons temporarily bound into assemblies are distinguished not only by an increase of their discharge rate, but also by the precise synchronization of their action potentials'' \cite{gray1989oscillatory, singer1995visual, singer1999neuronal}. 
They also predicted that neurons that respond to the same sensory object might fire in temporal synchrony, with a precision in the millisecond range. Synchronization by oscillation is briefly introduced in the following section.

\subsubsection{Synchronization by Oscillation}


\begin{figure}[ht]
    \centering
    \includegraphics[width=0.7\linewidth]{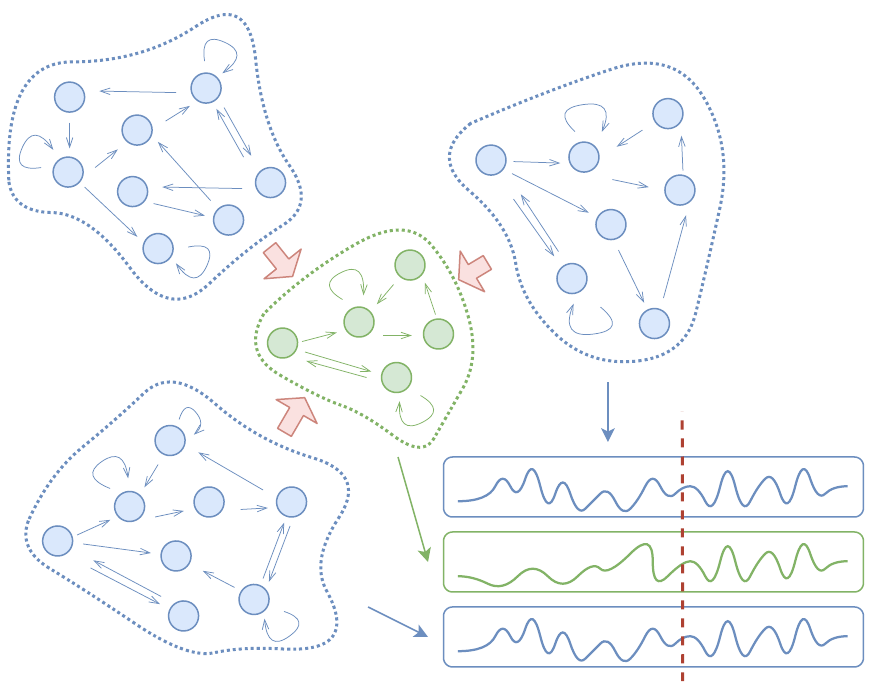}
    \caption{
    The spread of synchrony. Here, the blue neuron populations are already synchronized, and they might entrain other neurons (those in green) to become part of the same overall assembly after some point in time (see the dashed red line), thus resulting from the spread of synchronized activity through lateral connections. Figure modified from \cite{engel2001dynamic}.
    }
    \label{fig:fig_cog_spreadOfSync}
\end{figure} 

\begin{figure}[ht]
    \centering
    \includegraphics[width=0.5\linewidth]{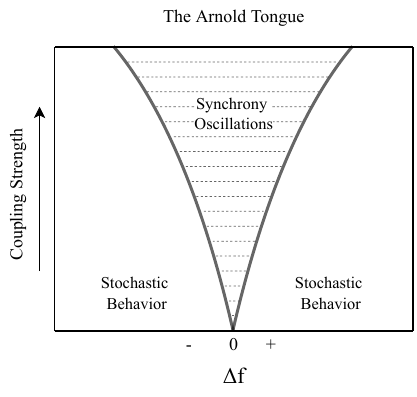}
    \caption{
    The Arnold tongue regime in coupled oscillators. This is a graphical representation modified from \cite{singer2018neuronal} to illustrate the synchronization behavior relating the difference in preferred frequency with increasing coupling strength, which results in a ‘tongue’-shape of possible synchronization regimes.
    }
    \label{fig:fig_cog_arnoldTongue}
\end{figure} 

\textbf{Definition of synchronization.}
The periodic changes of excitability of the neurons are considered neuronal oscillations among different brain areas \cite{singer2021cerebral}. It has been identified that these oscillations vary in terms of the frequency, ranging from approximately 0.05 Hz to 500 Hz \cite{buzsaki2004neuronal}.
According to \cite{fries2015rhythms}, synchronization by neuronal oscillations has been widely detected across various natural systems, in which different oscillations can ``coexist and often synchronized to each other or nested into each other''.
This observation is called the spread of synchrony. 
As an example shown in Fig. \ref{fig:fig_cog_spreadOfSync}, the synchronized larger-scale populations can entrain other smaller local assemblies with different oscillations to be overall synchronized. 
This spread of synchronized activity was then believed to be a reinterpretation of the represented objects \cite{engel2001dynamic}.

\textbf{Arnold tongue regime.}
One of the interesting observations of synchronization behavior in coupled oscillators is the Arnold tongue regime. 
An early experiment by Van Huygens revealed that the beats of pendulum clocks can be synchronized when having the same timber; that is, if the preferred frequencies of the oscillators are similar, weak mutual interactions are enough for oscillatory synchronization \cite{singer2018neuronal}.
This observation was then summarized as Arnold tongue regime by \cite{glass1994periodic}.
As shown in Fig. \ref{fig:fig_cog_arnoldTongue}, the coupling strength should increase in order to assure a stable synchronization when the preferred frequencies between coupled oscillators become increasingly different, thus resulting in a tongue-shaped pattern.
Synchronization would become unstable if the difference between preferred frequencies exceeds a critical point \cite{singer2018neuronal}.

\subsubsection{How RC Relates to Oscillations?}
Recall that in most of the existing literature, RC models should normally meet several requirements to be efficient and functional for various tasks.
These are also considered the properties of a RC model \cite{tanaka2019recent}, which include
(1) High-dimensionality: low-dimensional inputs are mapped into a high-dimensional space, which allows originally inseparable or temporal inputs to be linearly separable as shown by the Cover Theorem \cite{cover1965geometrical}.
(2) Non-linearity: non-linear mapping transforms the input into linearly separable reservoir states which can be read out by readout layer.
(3) Separation property: a RC model should be capable of separating different inputs into different classes, under small fluctuations or in noisy environments \cite{maass2002real}.
(4) Fading memory: also known as short-term memory or echo state property. This algebraic property eliminates the effect of initial network condition. In other words, it ensures that the reservoir state is dependent on recent-past inputs (reverberating responses), but not distant-past inputs (responses faded).

In a RC review, \cite{singer2021cerebral} made a proposal to link the concept of RC in machine learning and that of neuronal behaviors in the cognitive brain. Readers may refer to the article for details.
Based on the proposal, here we aim to discuss how the dynamics of coupled oscillators in mammalian brains could be exploited to accomplish the abovementioned four characteristics in RC.

\textbf{High-dimensionality and non-linearity.}
The first fact is that the cortex is reported to have consistent and random high-dimensional oscillations, which refers to the ``resting activity'' \cite{buzsaki2006Brain1}. 
\textcolor{black}{Meanwhile, it is believed that brains are likely to have an internal model of the external world (i.e., prior knowledge, which can be updated by learning).
}
When the input comes in, the input stimuli “activate” some feature-sensitive neurons, thus making the dynamics of the network collapse into a stimuli-specific substate (e.g., oscillatory synchronization). 
\textcolor{black}{
All of these may suggest that once a reservoir enters a substate (i.e., synchronization), it is likely that the dynamics can be tuned selectively to specific stimuli or generating specific output signals.
}

\begin{figure*}[ht]
    \centering
    \includegraphics[width=0.99\textwidth]{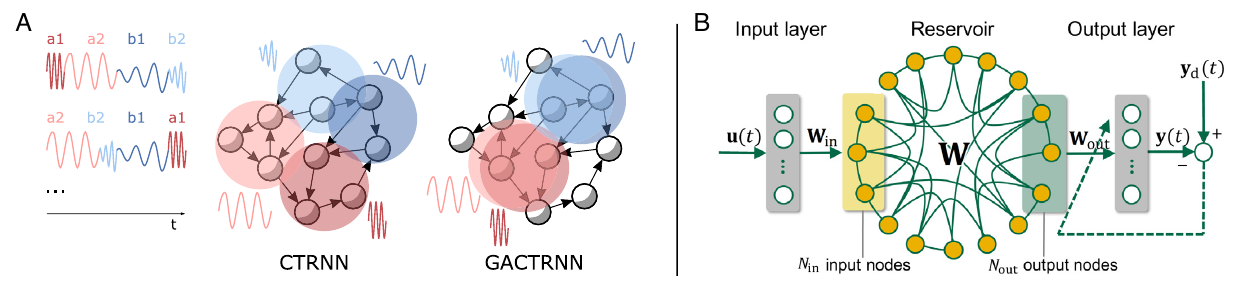}
    \caption{
    Neuroscientists suggest that neural oscillations reveal a highly complex interplay of neural populations and local integrations by coupled oscillations and synchronization, in which multiple timescales in hierarchical processing streams can be achieved. (A) and (B) show examples of how these mechanisms work on a computing and processing level.
    (A) GACTRNNs \cite{heinrich2020learning}, a type of neural network where neurons in a population can learn to simultaneously represent temporally different primitives, by changing their timescales.
    (B) An ESN with a small-world topology. The network rewiring probability is around 10\% \cite{kawai2019small}.
    } 
    \label{fig:TRNN and SW topology}
\end{figure*} 

\textbf{Separation property.}
As mentioned above, the stimuli-specific substate with rhythmic oscillations, according to \cite{singer2021cerebral}, ``would have a lower dimensionality and to exhibit less variance than the resting activity, to possess a specific correlation structure and be metastable due to reverberation (rhythmic oscillations) among nodes supporting the respective substate''. 
Note that all these activities, including the stimuli-specific substates, are happening in a high-dimensional state space; thus, different inputs can be well-separated and classified even linearly, similar to the readout of RC models.

\textbf{Fading memory.}
Moreover, fading memory refers to the short-term memory in RC literature, in which the reservoir state should depend on recent inputs but not distant-past inputs. 
From an oscillation point of view, there exist several experiments in the cat’s visual cortex, which can support the mechanism of fading memory  \cite{nikolic2009distributed}. 
As stated in \cite{singer2021cerebral}, these experiments show that ``
(1) the information about a particular stimulus persists in the activity of the network for up to a second after the end of the stimulus (fading memory).
(2) Two subsequent stimuli and the order of their presentation can be correctly classified with a linear classifier sometime after the end of the second stimulus, suggesting that the network is capable of performing non-linear XOR operations and
(3) Stimulus identity is distributed across many neurons ($>$30) and encoded both in the rate vector and the temporal correlation structure of the responses''.
The above evidence
may explain the fading memory from the perspective of neuronal oscillations, suggesting that short-term memory is not only a property of the networks, but also a consequence of the oscillations and reverberations.

In addition, long-term memory is also important, and it is a more complex one. 
Early experiments showed that the “default” state of the unperturbed, sleeping brain is a complex system of numerous self-governed oscillations, particularly in the thalamocortical system \cite{buzsaki2004neuronal, buzsaki2006Brain1}. 
The content of these oscillations reflects spike sequence patterns created by prior waking experience. 
Moreover, these oscillations are spontaneously replayed (e.g., during sleeping), leading to an “off-line” synaptic modification. Such replays might be the way to the formation of long-term memory.
Overall, a reason why short-term memory rather than long-term memory is one of the necessary requirements for building a RC model might be that we usually
keep the random connections in the reservoir fixed without modifications. As a result, RC models generally struggle to form long-term memory, since replays in terms of synaptic modifications are required (i.e., investigating long-term memory is a more challenging task).
In fact, some learning algorithms, such as STDP \cite{caporale2008spike} and FORCE learning \cite{sussillo2009generating}, are trying to modify the synaptic connections, and therefore they are likely to be able to possess long-term memory.

\subsubsection{Examples of Synchronization in RC}
\textbf{Multiple reservoirs.}
Deep reservoir computing was proposed in \cite{gallicchio2017deep}, in which multiple reservoirs are concatenated together to form a hierarchical network structure. Detailed model descriptions are discussed in Section \ref{Recent approaches in RC}. It has been proved that the deep RC structure can achieve
(1) multiple timescale representation, ordered along the network's hierarchy;
(2) multiple frequency representation, where progressively higher layers focus on progressively lower frequencies.
According to the Arnold tongue regime shown in Fig. \ref{fig:fig_cog_arnoldTongue}, if we keep the coupling strengths at a low level (week synaptic links, or even zero weight connections), neurons with similar preferred frequencies can be synchronized, at different frequency bands. Therefore, this may explain the reason why multiple reservoirs can achieve these while the conventional ESN cannot, from the perspective of neuronal oscillations.

\textbf{GACTRNN.}
Gating Adaptive Continuous Time Recurrent Neural Network (GACTRNN), is another research taking the inspirations from neuronal oscillation proposed in machine learning by \cite{heinrich2020learning} in 2020.
It extended the classic RNN to adaptive timescales RNN, which shares some similarities to reservoir computing models. 
GACTRNN is claimed to be able to learn to gate its timescale characteristic during activation and thus dynamically change the timescales in processing sequences; in other words, by changing their timescales during processing, neurons can learn to simultaneously represent temporally different primitives (Fig. \ref{fig:TRNN and SW topology}A).

\textbf{Small-world topology.} An ESN based on the topology of small-world (SW) wiring was proposed by \cite{kawai2019small} (Fig. \ref{fig:TRNN and SW topology}B). The model incorporated SW structure with RC and further investigated echo state property. It was found that the SW topology plays the roles of both efficient signal propagation and enhancement of the ESP in neural computation.
In fact, this idea partially originated from the cortical anatomical connectivity of the human brain.
According to \cite{buzsaki2004neuronal}, ``complex brains have developed specialized mechanisms for the grouping of principal cells into temporal coalitions''.
In order to reduce the complexity of the connections without excessive wiring, the number of long-range connections between neurons decreases in growing brains; in other words, the synaptic path lengths between distant cell assemblies are reduced, keeping the path lengths short and maintaining fundamental functions.



\section{Perspectives and Future Research} \label{perspectives}

Reservoir computing is becoming increasingly popular due to its simple network structure, hardware-friendly features, low computational cost, and fast training process.
These benefits enable RC to extend far beyond machine learning into a wide range of research fields.
In this paper, we provide a thorough overview of RC's history, strengths and weaknesses from the perspectives of machine learning, dynamical systems, physics, biology, and neuroscience. We also summarize recent advanced approaches and architectures for RC optimizations and implementations. 
Besides, applications of RC are reviewed, from which we have seen how this interdisciplinary idea can be applied in various research areas.

\textcolor{black}{
While RC still remains an unconventional computational framework compared to other machine learning techniques like deep learning, its impact can be enhanced by addressing various challenges. Recent developments have unveiled new directions and perspectives for RC, indicating its untapped potential and promising prospects that may even surpass those of mainstream methods. In this section, we present perspectives and discuss the open problems that motivate further research in this field.
}

\subsection{Reservoir Design and Optimization}
A consensus view of conventional RC is that initializing a random RNN as a reservoir is not the optimal solution, and that connecting a linear readout with the reservoir limits the generation of the downstream responses. 
It is also known that neurons in cortical networks in the brain are not randomly connected, while their structures and synapses exploit an evolutionary and developmental process \cite{subramoney2021reservoirs}.
Recent research, especially on ESNs and LSM, mainly focuses on network structure designs (e.g., deep reservoir), parameter optimizations (e.g., particle swarm optimization) and training rule determinations (e.g., STDP and Hebbian learning).
Even if the optimal synaptic weights were discovered, the performance of various concrete tasks would still vary.
As pointed out by Jaeger \cite{nakajima2021reservoir}, ``currently available insights are mostly distilled from experimental studies of timescale profiles or frequency spectra in input data and provide no comprehensive guides for optimizing reservoir designs''.
In other words, one should find a way to analyze and abstract both the characteristics of input/output and task specifications, which can be used to design the reservoir dynamics.
One possible solution is called reservoir Learning-to-learn (L2L) \cite{subramoney2021reservoirs}, in which a set of (hyper)parameters of the reservoir are optimized by BPTT for a whole family of learning tasks; note that this shares some similarity to meta learning in machine learning and neuroscience. 
This L2L method was investigated on LSM models and it can also be implemented by other RC architectures. 
\textcolor{black}{Moreover, to have better and faster learning, it is possible to train the reservoir by the L2L method even without changing synaptic weights to readout neurons.
Nevertheless, whether the performance takes advantage of other reservoirs is still an open question.}

\textcolor{black}{
\subsection{Easy-access Tools, Coding frameworks and Recipes}
One pressing open problem in RC is the relative lack of user-friendly coding environments, libraries, and computation frameworks. 
Unlike the well-developed infrastructure supporting deep learning and other scientific computational paradigms, the coding ecosystem for RC remains relatively underdeveloped and fragmented. 
As shown in this paper that although numerous models and architectures have been proposed for RC, there is a notable dearth of unified frameworks that researchers can leverage with ease. 
This poses a significant challenge as it lowers the speed and efficiency of research, requiring additional time and effort to navigate through a variety of individual tools and frameworks, and often necessitates the development of custom code for each research project. 
}

\textcolor{black}{
However, this does not imply that no progress is being made towards building a more unified and accessible coding infrastructure for RC. 
Indeed, we have seen some promising developments over the years.
Back in the early 2010s, \cite{lukovsevivcius2012practical} provided a comprehensive guide on how to implement an ESN, which widely impacts future studies. 
The author also released a demonstration of coding ESN from scratch in Julia, Matlab, Octave, Python, and R language. 
In 2012, a toolbox was developed for RC called \textit{Oger} (OrGanic Environment for Reservoir computing) to train and evaluate recurrent neural networks, particularly ESNs and LSMs \cite{verstraeten2012oger, oger2012}.
From 2017 on, several tools were released in terms of different RC models, such as ESNs, LSMs, and FORCE-based algorithms.
The \textit{easyesn} library was released \cite{easyesn2017, thiede2017easyesn}, providing a more easy-to-use API for automatic gradient based hyperparameter tuning (of ridge regression penalty, spectral radius, leaking rate and feedback scaling), as well as transient time estimation. 
Meanwhile, a hands-on LSM implementation using \textit{NEST} simulator in Python was proposed \cite{kaiser2017scaling, gewaltig2007NEST}, which is considered to be the starting point for new researchers who are interested in spiking-based RC models.
Later, another open-source spiking model framework, \textit{Nengo}, was developed for FORCE learning and its variation implementations \cite{bekolay2014nengo}.
In 2018, \textit{EchoTorch} was proposed, and perhaps it is the first Python package to simplify the evaluation and implementation of ESNs and RC \cite{echotorch2018}.
In 2019, a Matlab toolbox for DeepESNs \cite{gallicchio2017deep} was released that extends the RC paradigm towards deep networks. 
One of the most common deep learning frameworks, TensorFlow, also supports the ESN layer in 2020 (see TF Addons). 
More recently, in 2022, \cite{liu2022tension} present \textit{tension}, an object-oriented, open-source Python package that implements a TensorFlow / Keras API for FORCE learning. 
Another Julia package for RC is \textit{ReservoirComputing.jl} \cite{martinuzzi2022reservoircomputing}. It aims to provide a simple and flexible framework to work with ESNs and other models.
Additionally, \cite{trouvain2022create} present a Python library that facilitates the creation of RC architectures, from ESNs and FORCE learning, to complex networks such as DeepESNs and other advanced architectures with complex connectivity between multiple reservoirs with feedback loops.
}

\textcolor{black}{
Despite progress, the full potential of RC is yet to be realized, and the goal of a unified and accessible environment for RC still eludes us. There is a need for more work in this area to ensure that the full potential of RC can be explored and utilized effectively.
}

\begin{figure*}[ht]
    \centering
    \includegraphics[width=0.99\textwidth]{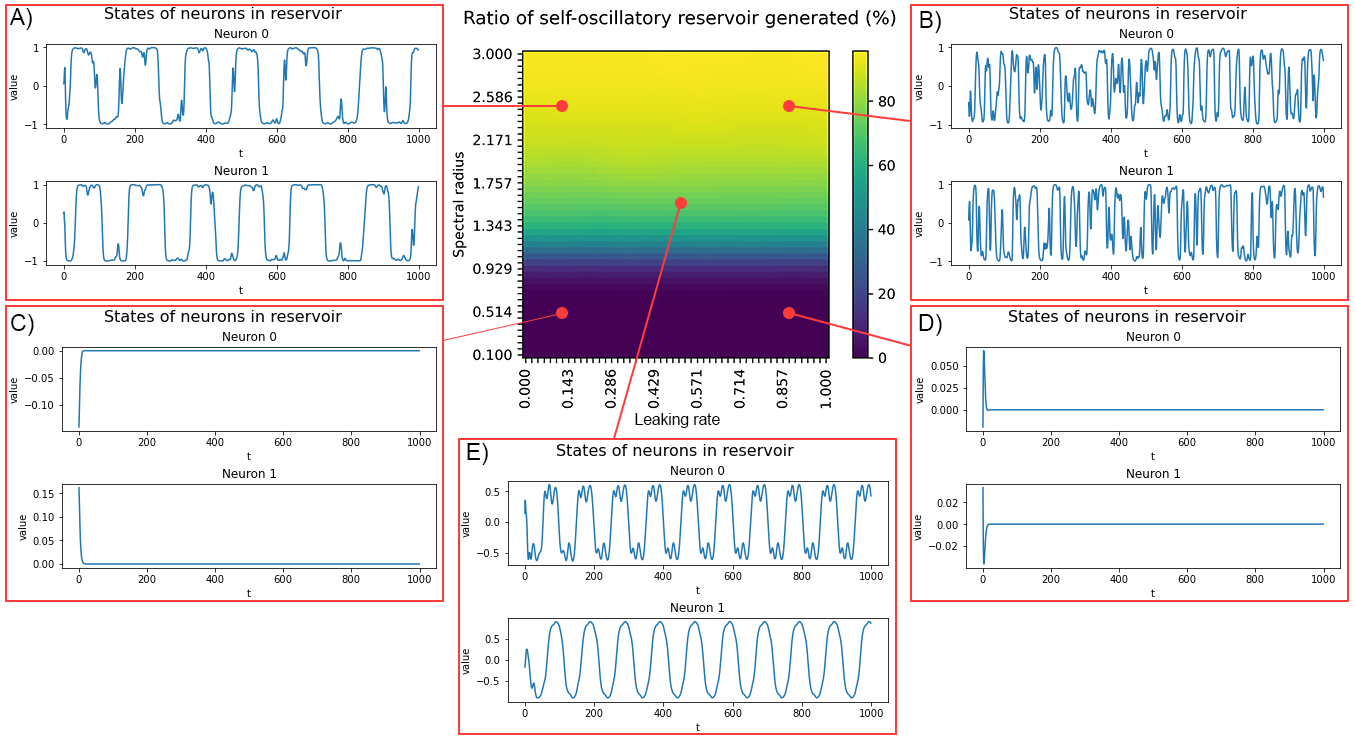}
    \caption{
    \textcolor{black}{Self-oscillatory echo state network (SO-ESN) generating oscillations to reproduce desired waveforms without any inputs (figure from \cite{foong2023generating}, with permission). The figure shows the \textit{ratio} (\%) of reservoir that triggers oscillations with regard to leaking rate (x-axis) and spectral radius (y-axis).  A stable synchronization by oscillation can be seen in (E), in which neurons 0 and 1 were randomly selected from the population. 
    }
    } 
    \label{fig:soesn}
\end{figure*}

\subsection{Physical RC and Extremely Efficient Hardware}
As mentioned earlier, RC has reawakened and gained attention because of the fast development of PRC designs.
Unlike conventional RCs that suffer from the information processing speed limit, PRCs can overcome this limit and process massive amounts of data in real-time. 
We review PRC models that use different physical materials from different areas such as electronics, optics, chemistry, and quantum. 
As more and more PRC approaches are being proposed, there are several open problems to be solved. 
For example, some PRCs get rid of the massive recurrent connections in RNNs, yet they are hard to design and tune (e.g., implementing a delayed feedback loop of the single node reservoir is quite a challenging task). 
Similarly, setting hyperparameters for PRC is not straightforward.
In 2021, \cite{dale2021reservoir} proposed a framework to evaluate what makes a good computing substrate, providing a new perspective on how to build and compare physical reservoir computers.

\textcolor{black}{
Another challenge lies in harnessing the full potential of RC for machine learning applications and achieving highly efficient hardware implementations. Very recently, this issue has been explored and discussed in \cite{tanaka2022guest}, where the future directions of inquiry are segmented into three categories.
Simply put, the first category focuses on the theoretical aspects of RC, aiming to drive efficient model design and ensure reliable RC applications. This includes works like reservoir memory machines \cite{paassen2022reservoir}, consistency capacity \cite{jungling2022consistency} and curve fitting abilities \cite{manjunath2022echo} analysis.
The second category delves into the exploration of novel model designs and applications of RC, with an aim to enhance computational performance and efficiency in tasks related to pattern recognition. Possible solutions include (1) industrial applications such as adaptive practical nonlinear model predictive control \cite{schwedersky2022adaptive} and digital twins \cite{kong2023reservoir}; (2) integrating RC with deep learning methods such as convolutional and graph neural networks \cite{jalalvand2022real, pasa2022multiresolution, pedrelli2022hierarchical, tong2018reservoir, tanaka2022reservoir}.
Lastly, the third category is to keep investigating new architectures and mechanisms in physical hardware that are suitable for RC implementations. Latest research includes a new FPGA-based RC for low-power pattern recognition \cite{gupta2022neuromorphic}, networks based on the Schrödinger equation \cite{nakajima2022neural}, and a new cellular automata implementing rule (CA90) \cite{kleyko2022cellular}.
Overall, these studies use a wide range of new hardware, showcasing efficient RC-based methods and stimulating further growth in this research domain.
}

\subsection{RC with Cognitive Science and Neuroscience}
In terms of modelling RC in cognitive science and neuroscience, we have reviewed several mechanisms of cognitive brains and tried to bridge the concept of neuron population to RC, such as \textit{mixed selectivity} \cite{rigotti2013importance} which reveals that the prefrontal cortex can generate high-dimensional mixed selective dynamics to assure the separability in the downstream readout units.
Future research directions include finding new analogies for biological characteristics in RC, allowing for a deeper understanding of brain's and bodies' mechanisms.

\textcolor{black}{
The domain of oscillation with synchronization, as previously discussed, is one area warranting further investigation. 
In fact, progress has been made in using ESNs to produce oscillatory outputs without any inputs, by mimicking the central pattern generators (CPGs) shown to be involved in rhythmic human movement \cite{cpg1, cpg2, cpg3, cpg4}.
CPGs are important circuits present in the neural system of live beings. In fact, vertebrates have a spinal cord composed of many of CPG circuits, and it was shown that the spinal cord and mostly CPGs are sufficient for complex locomotion (e.g., walking and running in cats) \cite{yuste2005cortex, duysens1998neural}.
In 2023, Tham and Vargas \cite{foong2023generating} show that even the most basic ESN can be trained to reproduce the trajectory of dynamical systems from simple sinusoidal and square waves to complex Lorenz chaotic time series with high precision, without any external excitation. }
\textcolor{black}{
Fig. \ref{fig:soesn} depicts the different probabilities of having a reservoir that triggers oscillations in terms of both leaking rate and spectral radius (i.e., the echo state property, ESP), where pure yellow represents the highest probability. 
Here, it is important to note that the ESP, by design, typically restricts spontaneous oscillation, as it ensures that the reservoir’s internal state should eventually lose memory of its initial conditions, hence creating a fading ``echo'' of past inputs. 
This can be seen in the cases of Fig. \ref{fig:soesn}C-D, that the reservoirs' states converge to 0 and remain stable afterward (refer to  damped oscillation in \cite{foong2023generating}). 
In contrast, as shown in Fig. \ref{fig:soesn}E, oscillations occur when the ESP is most likely not to be guaranteed.
Therefore, this study clearly indicates the necessity for additional research, particularly regarding the outcomes and implications when the ESP is not strictly adhered to.
}

On the other hand, high-dimensionality is another feature shared by the encoding of neuron populations and RC models. 
Recent studies have also proposed other approaches to understanding the role of high-dimensionality in biological and artificial neural networks.
Neural population geometry, for example, is an approach that provides a useful population-level mechanistic descriptor underlying task implementation.
Here, the geometry of representation can be represented by high-dimensional neural activity, and it is further observed that the neural activity lies on lower-dimensional subspaces, i.e., the so-called intrinsic dimensionality or neural manifolds \cite{jazayeri2021interpreting}.
Similar to the reservoir responses, these lower-dimensional subspaces can then be well-separated by using simple classifiers.

Although all of these remain in an early stage of development,
the investigations look promising for getting deeper insights into both artificial and natural intelligence.


\subsection{RC from an Evolutionary Perspective}
In addition to artificial intelligence and neuroscience, some researchers are exploring RC through an evolutionary lens, which brings a fresh angle to the discussion.
Natural systems exhibiting reservoir-like behaviors are prevalent in nature (e.g., nonlinearities in liquids \cite{maass2002real}, soft robotic in muscles \cite{nakajima2013soft}, electric and chemical dynamics in neural networks \cite{maass2004methods, nguyen2020reservoir}, and brain mechanisms \cite{dominey1995complex, dominey1995model, rigotti2013importance, singer2021cerebral}). 
This suggests that such reservoirs might have \textit{evolved} as advantageous structures for processing complex information, similarly to other computational approaches such as feed-forward networks \cite{lecun2015deep}, attractor networks \cite{hopfield1982neural}, and self-organized maps \cite{kohonen1982self} that have been influenced by and have influenced biology. 
As stated by \cite{seoane2019evolutionary}, however, the current evidence is not as compelling as the well-established similarities between, for example, the structure of the human visual system and deep convolutional neural networks. 
From an evolutionary perspective, \cite{seoane2019evolutionary} argue that although reservoir-like systems may initially emerge due to their simplicity, their long-term persistence could be hindered by evolutionary trends towards specialization and scaling. 
In other word, as the reservoir evolves to specialize, integrate diverse sensory information, or scale up, the generalizing properties of the reservoir may become less advantageous compared to highly specialized circuits.
This leads to another question: under what conditions might reservoirs maintain their original architecture with redundant dynamics, and when might they evolve towards more specialized configurations? 
Overall, understanding the evolutionary constraints is essential in evaluating its potential and limitations both in biological systems and engineering applications.

\subsection{Hybrids and New Foundations}
Last but not least, some recent works from 2019 have merged RC with other systems or paradigms such as deep learning \cite{gallicchio2017deep, long2019evolving, challita2019interference, zhou2020deep, li2022multi, liu2022reservoir, jalalvand2022real, pasa2022multiresolution, pedrelli2022hierarchical, tong2018reservoir, tanaka2022reservoir}. Moreover, some authors have started investigating modifying the foundations (e.g., NG-RC \cite{gauthier2021next}).
In fact, research on random networks \cite{carroll2019network, kawai2019small, cucchi2021reservoir, berner2021patterns} shows that there is a wide range of methods taking inspiration from some core mechanics of RC to build novel approaches that might challenge reservoir computing's current foundations in the near future.

\section*{Acknowledgment}
This work was supported by JSPS Grant-in-Aid for Challenging Exploratory Research—Grant Number JP22534665, JST Strategic Basic Research Promotion Program (AIP Accelerated Research)—Grant Number 22584686, JSPS Research on Academic Transformation Areas (A)—Grant Number 22572551, JST SPRING—Grant Number JPMJSP2136.

\bibliographystyle{unsrt}
\bibliography{main_arxiv.bbl}

\end{document}